\documentclass[10pt,twocolumn,letterpaper]{article}

\usepackage[pagenumbers]{iccv} 

\definecolor{iccvblue}{rgb}{0.21,0.49,0.74}
\usepackage[pagebackref,breaklinks,colorlinks,allcolors=iccvblue]{hyperref}

\usepackage{graphicx}
\usepackage{amsmath}
\usepackage{amssymb}
\usepackage{booktabs}
\usepackage{multirow}
\usepackage{bm}
\usepackage{comment}
\usepackage{color, colortbl}
\usepackage{hhline}
\definecolor{Gray}{gray}{0.96}
\newcolumntype{g}{>{\columncolor{Gray}}c}
\usepackage{here}
\usepackage{pifont}
\newcommand{\xmark}{\ding{55}} 
\definecolor{mygray}{rgb}{0.2,0.2,0.2}

\def\paper{RealTraj}

\def\paperID{*****} 
\def\confName{ICCV}
\def\confYear{2025}

\title{RealTraj: Towards Real-World Pedestrian Trajectory Forecasting}

\author{Ryo Fujii$^{1}$ \hspace{0.6cm} Hideo Saito$^{1}$ \hspace{0.6cm} Ryo Hachiuma$^{2}$ \\
$^{1}$Keio University \hspace{0.6cm} $^{2}$NVIDIA \\
{\tt\small \{ryo.fujii0112, hs\}@keio.jp, rhachiuma@nvidia.com}\\
}

\begin{document}
\maketitle
\begin{abstract}
This paper jointly addresses three key limitations in conventional pedestrian trajectory forecasting: pedestrian perception errors, real-world data collection costs, and person ID annotation costs. We propose a novel framework, \textit{\paper}, that enhances the real-world applicability of trajectory forecasting. Our approach includes two training phases—self-supervised pretraining on synthetic data and weakly-supervised fine-tuning with limited real-world data—to minimize data collection efforts. To improve robustness to real-world errors, we focus on both model design and training objectives. Specifically, we present Det2TrajFormer, a trajectory forecasting model that remains invariant to tracking noise by using past detections as inputs. Additionally, we pretrain the model using multiple pretext tasks, which enhance robustness and improve forecasting performance based solely on detection data. Unlike previous trajectory forecasting methods, our approach fine-tunes the model using only ground-truth detections, reducing the need for costly person ID annotations. In the experiments, we comprehensively verify the effectiveness of the proposed method against the limitations, and the method outperforms state-of-the-art trajectory forecasting methods on multiple datasets. The code will be released at
\href{https://fujiry0.github.io/RealTraj-project-page}{project page}.
\end{abstract}

\section{Introduction}
\label{sec:intro}
\begin{figure}[tb] 
\centering
\includegraphics[width=0.9\linewidth]{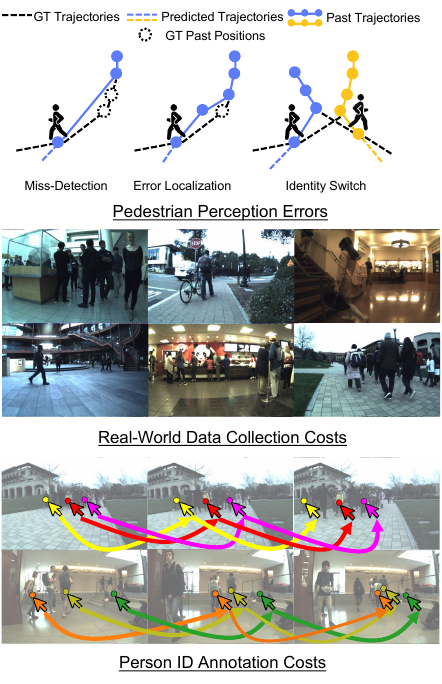}
\caption{Our paper addresses the three limitations in the existing pedestrian trajectory forecasting task. (Top) Pedestrian perception errors can significantly degrade trajectory forecasting performance. (Middle) Real-world data collection necessitates substantial manual effort. (Bottom) Person ID annotations require extensive manual labor.}
\label{fig:motivation}
 \vspace{-1.5em}
\end{figure}

The pedestrian trajectory forecasting task aims to predict the future positions of pedestrians based on their past observed states, a critical capability for understanding motion and behavioral patterns~\cite{Rudenko2019HumanMT}. This task plays a crucial role in real-world applications such as social robotics navigation~\cite{jetchev2009robot, foka2010robotnavi}, autonomous driving~\cite{cui2019autonomous}, and surveillance systems~\cite{omar2021surveillance}. Over the years, deep learning models have dominated this domain with the various model architectures, such as RNNs~\cite{alahi206sociallstm, gupta2018socialgan, kothari2022human, Bae_2024_CVPR}, graphs~\cite{bae2023graphtern, salzman2020trajectronplusplus, bae2022gpgraph, xu2023eqmotion}, Transformers~\cite{giuliari2021transformer, girgis2022autobot, shi2023tutr, saadatnejad2024socialtransmotion} and 
diffusions~\cite{mao2023led,gu2022mid,bae2024singulartrajectory,han2024GeoTDM}, and these models are evaluated on large-scale benchmarks~\cite{zhou2012gcs, pellegrini2010eth, leal2014ucy, robicquet2016sdd, bock2020ind, zhan2018generating, martin2021jrdb}. However, existing approaches are evaluated under fixed, well-known benchmarks with ideal experimental conditions, such as completely annotated trajectories with the pre-recorded videos. This limits their applicability to real-world scenarios, necessitating further performance enhancements for practical deployment. In this paper, we address three key limitations present in existing approaches, as described below (\cref{fig:motivation}).

\noindent \textbf{Pedestrian Perception Errors.} 
Conventional approaches often assume that each pedestrian's past trajectory is perfectly obtained through upstream perception modules: pedestrian detection and tracking. However, this assumption is unrealistic in real-world scenarios, where obtaining complete trajectories is challenging due to detection errors (\eg, miss-detections and localization errors) and tracking errors (\eg, identity switches). When trajectory forecasting models use imperfect trajectories as input, their performance will degrade significantly~\cite{yu2021towardrobust, weng2022mtp, weng2022whosetrack, xu2024realworldforecasting, fujii2025crowdmac}.

\noindent \textbf{Real-World Data Collection Costs.} Conventional trajectory forecasting approaches are typically trained on large-scale datasets and evaluated on data domains similar to the training (\eg, consistent camera angles and sensor setups). However, these methods often overlook the substantial costs associated with real-world data collection, including the acquisition of raw sensor data such as point clouds or videos, which requires significant manual effort. Thus, it is crucial to develop prediction models that can be trained effectively while minimizing real-world data collection efforts.

\noindent \textbf{Person ID Annotation Costs.} 
In addition to the challenge of large-scale real-world data collection, most existing trajectory forecasting models rely on fully supervised training. This approach necessitates ground-truth future trajectories, including precise pedestrian positions and consistent person identities across frames.

In this paper, we propose a novel trajectory forecasting framework, \textit{\paper}, designed to enhance the real-world applicability of trajectory forecasting by jointly addressing aforementioned three limitations. Our approach consists of two training phases—self-supervised pretraining and weakly-supervised fine-tuning—followed by an inference phase. First, we leverage large-scale synthetic trajectory data generated from a simulator~\cite{berg2011ocra} to effectively learn trajectory patterns using only synthetic data, reducing the need for extensive real-world data collection (addressing the second limitation). However, relying solely on clean synthetic data weakens robustness against real-world errors. 

To mitigate this, we enhance both the model design and training objectives: we introduce \textit{Det2TrajFormer}, a trajectory forecasting model designed to be invariant to tracking noise by only inputs past \textit{detections} (without any person indices across frames). Additionally, inspired by the recent progress of self-supervised learning mechanism~\cite{he2022mae,wu2023dmae,Xu_2023_ICCV}, we pretrain the model with multiple self-supervised pretext tasks, such as unmasking, denoising, and person identity reconstruction, to improve robustness against perception errors (addressing the first limitation) while enhancing forecasting performance solely from detections. 

Finally, unlike previous trajectory forecasting approaches, we propose training the model exclusively on ground-truth detections during the fine-tuning phase, thereby reducing person ID annotation costs (addressing the third limitation) on the real data. We further introduce an acceleration regularization term to discourage abrupt changes in acceleration, resulting in smoother and more realistic trajectory forecastings. During inference, Det2TrajFormer takes detection results as input to forecast future trajectories.

In summary, our contributions are three-fold: (1) We introduce a novel trajectory forecasting framework, \textit{\paper}, with self-supervised pretraining on large-scale synthetic data and weakly-supervised fine-tuning on limited real-world data, enhancing real-world applicability by jointly addressing pedestrian perception errors, data collection costs, and person ID annotation costs within a unified and single framework. (2) We propose three self-supervised pretext tasks to improve robustness against pedestrian perception errors and enhance trajectory forecasting accuracy using only detections. (3) We reduce person ID annotation costs by training the model solely on ground-truth detections via proposed weakly-supervised loss.

\section{Related Work}
\subsection{Robust Trajectory Forecasting}
Pedestrian trajectory forecasting models aim to predict future positions based on observed trajectories. Deep learning methods demonstrating strong performance due to their representational capabilities~\cite{alahi206sociallstm, gupta2018socialgan, mohamed2020sstgcnn, xu2022groupnet, kosaraju2019sbigat, mangalam2021ynet, giuliari2021transformer, sadeghian2019sophie, ivanovic2019trajectron, mangalam2020pecnet, salzman2020trajectronplusplus, gu2022mid, xu2022memonet, mao2023led, saadatnejad2024socialtransmotion, xu2023eqmotion, girgis2022autobot,han2024GeoTDM,Damirchi_2025_WACV}. 

Despite the significant advancements, most trajectory forecasting models operate under idealized conditions, relying on ground-truth past trajectories for training and evaluation while ignoring the impact of imperfect inputs. Recent studies have increasingly recognized the challenges posed by perception module errors~\cite{weng2022mtp, fujii2021two, xu2023uncovering, zhang2023fromdetection, zhang2024oostraj, chib2024mstip, weng2022whosetrack, xu2024realworldforecasting, xu2024flexilength, pranav2024missingbenchmark, feng2023macformer,sun2022momentary,Li2023BCDiff,Li2024LaKD} and adversarial attacks~\cite{zhang2023trajpac, cao2022advdo, zhang2022adversarial, jiao2023semiadv}. This work focuses on enhancing robustness against perception module errors, which are unavoidable in real-world deployments and can significantly degrade prediction accuracy~\cite{yu2021towardrobust, weng2022mtp, weng2022whosetrack, xu2024realworldforecasting, fujii2025crowdmac}. Some approaches address incomplete observations, encompassing missing data~\cite{fujii2021two, xu2023uncovering,chib2024mstip,pranav2024missingbenchmark}, momentary visibility~\cite{sun2022momentary,Li2023BCDiff,monti2022howmany}, variations in observation length~\cite{xu2024flexilength,Li2024LaKD}, and out-of-sight instances~\cite{zhang2024oostraj} caused by occlusions or limited viewpoints (\ie, detection errors). Others focus on mitigating  tracking errors~\cite{yu2021towardrobust, weng2022mtp, weng2022whosetrack, zhang2023fromdetection, feng2023macformer}. Unlike the previous approaches, which independently tailored the approach for the specific error types, we aim to improve the robustness jointly against both detection (miss and wrong detection) and tracking errors within a single framework.

\begin{figure*}[tb] 
\centering
\includegraphics[width=\linewidth]{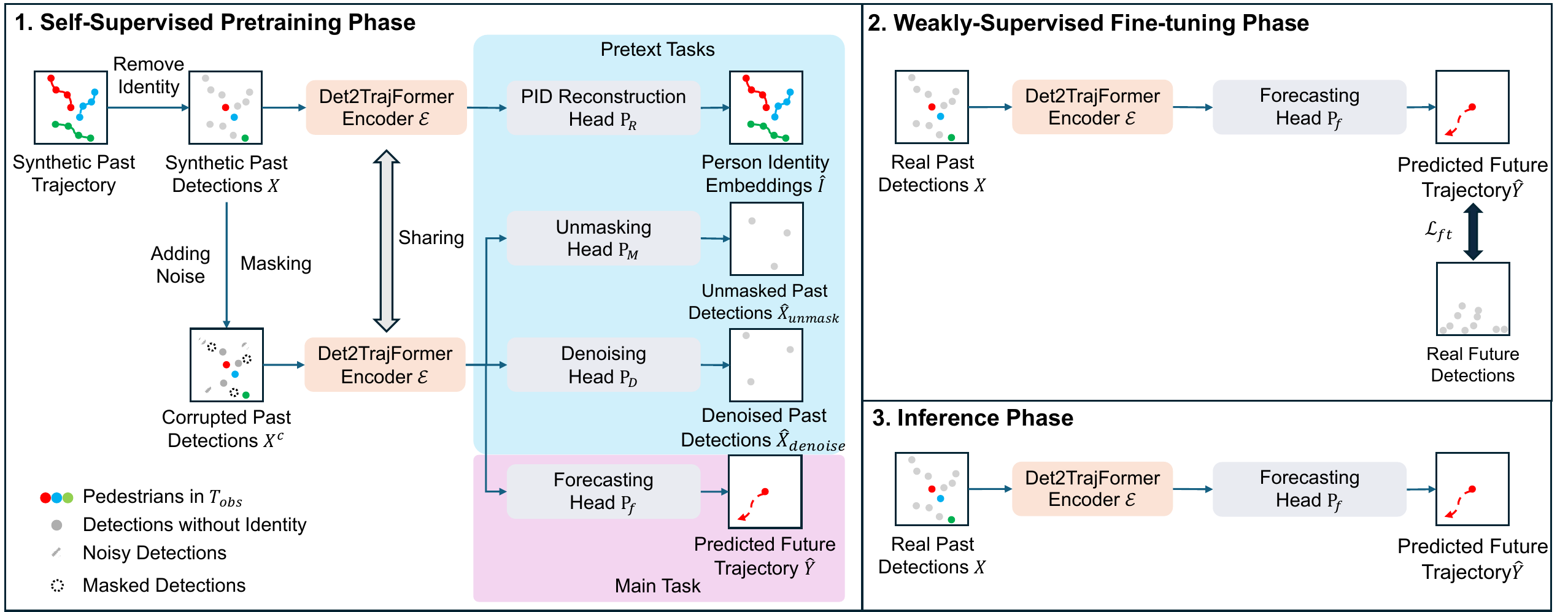}
\caption{Our proposed framework consists of two training phases and an inference phase. (1) Self-supervised pretraining on synthetic data using multiple pretext tasks. (2) weakly-supervised fine-tuning on real ground-truth detections. (3) Future trajectory inference based solely on detection inputs.}
\label{fig:overview}
\vspace{-1.em}
\end{figure*}

\subsection{Adapting Trajectory Forecasting}
Conventional trajectory forecasting models often heavily depend on specific training data domains, overlooking the substantial costs and manual effort required for real-world data collection, such as collecting the videos to obtain the trajectories. To address this challenge, recent research has focused on lightweight methods for adapting pretrained prediction models to newly captured data. Some approaches target cross-domain transfer~\cite{xu2022adaptive, ivanovic2023expanding, li2023syntheticdriving}, while others emphasize online adaptation~\cite{park202t4p, li2022online, cheng2019human}, continual learning~\cite{marchetti2020mantra, xu2022memonet, sun2021three}, and prompting-based strategies~\cite{thakkar2024adaptive}. These methods have demonstrated notable improvements in forecasting performance. However, they often assume access to ground-truth trajectories, which require costly person ID annotations. To reduce this burden, we propose to fine-tune the model using only ground-truth detections, thereby eliminating the extensive need for person ID annotations.

\subsection{Self-Supervised Learning}
Self-supervised learning (SSL) is a promising approach that enables models to learn valuable latent features from unlabeled data. Through pretraining on pretext tasks and pseudo-labels derived from data, followed by fine-tuning on downstream tasks, SSL has facilitated significant advancements in computer vision~\cite{he2022mae} and natural language processing (NLP)~\cite{devlin-etal-2019-bert}. However, few studies have explored SSL in trajectory forecasting~\cite{xu2022pretram, bhattacharyya2022ssllanes, cheng2023forecastmae, chen2023trajmae, mozghan2024dyset, li2023syntheticdriving, gao2024multitransmotion}. Applying SSL to trajectory forecasting poses unique challenges, as it typically requires large-scale annotated data for pretraining. Unlike fields such as computer vision and NLP, which benefit from abundant unlabeled data, trajectory forecasting relies on annotated trajectories—often involving costly sensor setups and extensive human annotation—limiting scalability and the potential of SSL. Inspired by advancements in training with synthetic data in image domains~\cite{kataokaijcvfractaldb, mu2020syntheticanimals, dosovitskiy2015flownet, teed2020eccv}, we explore self-supervised pretraining using synthetic data to reduce data collection and annotation costs. Additionally, we introduce multiple pretext tasks designed to enhance model robustness against pedestrian perception errors and improve trajectory forecasting performance from detection inputs.

\section{\paper}
\subsection{Problem Formulation}
We aim to forecast the future trajectory of a target pedestrian in the scene based on detections from observed frames, including the non-targeted pedestrians in the environment. Unlike prior work, our approach skips the tracking step and instead uses detections directly, thereby avoiding the propagation of tracking errors into the forecasting task. Formally, let $Y = (y_1, y_2, \dots, y_{T_{pred}}) \in \mathbb{R}^{T_{pred} \times 2}$ denote the future trajectory of the target pedestrian over $T_{pred}$ time steps where each position is represented as a tuple $y_t=(u_t, v_t) \in \mathbb{R}^{2}$, and let $X = (X_1, X_2, \dots, X_{T_{obs}}) \in \mathbb{R}^{KT_{obs} \times 2} $ represent the set of past detections of $K$ pedestrians over $T_{obs}$ time steps. The detections at time $t$ are defined as $X_t = (x_t^1, x_t^2, \dots, x_t^{K}) \in \mathbb{R}^{K \times 2}$, where each detection is the position of pedestrian $x_t^{k}=(u_t^{k}, v_t^{k}) \in \mathbb{R}^{2}$ at timestep $t$\footnote{Unlike the conventional trajectory forecasting approaches, our method takes a set of detections as input at each frame. Therefore, it is not guaranteed that $x_t^1$ and $x_{t+1}^1$ correspond to detections of the same pedestrian.}. The target pedestrian is one of the $K$ pedestrians in the last observed frame ($t = T_{obs}$). The model can be applied $K$ times to predict $K$ pedestrians' trajectories. The target pedestrian is specified by translating the detections to make one of the detections in the last frame to the origin. The goal is to learn a function $\mathcal{F}(X)$ that maps the input detections $X$ to a predicted future trajectory $Y$ of a target pedestrian, such that $Y = \mathcal{F}(X)$.

\subsection{Overview}
Our framework consists of two training phases—self-supervised pretraining and weakly-supervised fine-tuning—along with an inference phase, as illustrated in \cref{fig:overview}. We leverage large-scale synthetic trajectory data from a simulator~\cite{berg2011ocra} to effectively learn patterns while reducing the need for extensive real-world data collection. To ensure real-world robustness, we approach from both the training scheme and the trajectory forecasting model $\mathcal{F}$, Det2TrajFormer. During the pretraining, we leverage multiple pretext tasks to improve robustness and forecasting accuracy using large-scale synthetic trajectory data. During the fine-tuning, the model is fine-tuned solely on real-world ground-truth detections, eliminating the need for costly person ID annotations. During inference, Det2TrajFormer predicts future trajectories from observed detections.

\paragraph{Det2TrajFormer.}
Det2TrajFormer consists of an embedding layer $\mathcal{G} \in \mathbb{R}^{2 \rightarrow d}$, Transformer encoder~\cite{Vaswani2017Neurips} $\mathcal{E} \in \mathbb{R}^{d \rightarrow d}$, and a forecasting head $\mathcal{P}_f \in \mathbb{R}^{d \rightarrow 2}$. Each past detection $x$ is input to the embedding layer to obtain $d$-dimensional embedding along with sinusoidal time index encoding~\cite{Vaswani2017Neurips}. We concatenate $\mathcal{G}(X)$ with $T_{pred}$ learnable query tokens ($Q \in \mathbb{R}^{T_{pred} \times d}$) and input them into $\mathcal{E}$ to obtain $\hat{H}$ and $\hat{Q}$. $\hat{H} \in \mathbb{R}^{KT_{obs} \times d}$ are output tokens corresponding to the input tokens, $\mathcal{G}(X)$. Finally, the future trajectories $\hat{Y}$ can be predicted from $\hat{Q}$, as specified as follows:
\begin{align}
\label{eq:main-trajectory-prediction}
\begin{split}
[\hat{H}, \hat{Q}] &= \mathcal{E}([\mathcal{G}(X), Q]), \\
\hat{Y} &= \mathcal{P}_{f}(\hat{Q}),
\end{split}
\end{align}
where $[\cdot]$ denotes the concatenation operation along the token axis. Following recent works~\cite{gao2024multitransmotion,cheng2023forecastmae}, our approach can incorporate multiple forecasting heads ($\mathcal{P}_f^{n}$) to generate \( N \) possible future predictions: 
\(\hat{Y}^{n} = (\hat{y}_1^{n}, \hat{y}_2^{n}, \dots, \hat{y}_{T_{pred}}^{n}) \in \mathbb{R}^{T_{pred} \times 2}, \quad n = 1, 2, \dots, N\). For simplicity, we describe our method using a single future prediction setting.

\subsection{Self-Supervised Pretraining Phase}
Using synthetically generated trajectories, we train the model on the primary forecasting task while incorporating pretext tasks to enhance performance and improve robustness against detection errors.
From the synthetically generated trajectories, we generate a set of past detections $X$ by removing the identity information. Additionally, we simulate the miss-detection and localization errors to $X$ by randomly masking and adding Gaussian noises, resulting in the corrupted past detections, $X^C$. We propose three pretext tasks that aim to reconstruct the masked detections from $X^C$ (unmasking), denoise the $X^C$ (denoising), and reconstruct the removed person ID information from $X$ (person ID reconstruction).
Only during the pretraining, we additionally employ three heads  $\mathcal{P}_{M} \in \mathbb{R}^{d \rightarrow 2}$, $\mathcal{P}_{D} \in \mathbb{R}^{d \rightarrow 2}$, and $\mathcal{P}_{R} \in \mathbb{R}^{d \rightarrow d_I}$, where $d_I$ denotes the dimension of person ID embedding.

\noindent \textbf{Trajectory Forecasting Task.}
The trajectory forecasting task aims to forecast the future trajectory from detections. To improve robustness against detection errors during inference, we forecast the future trajectory $\hat{Y}$ from corrupted detections $X^C$ using \cref{eq:main-trajectory-prediction}.

\noindent \textbf{Unmasking Task.} 
The goal of the unmasking pretext task is to encourage the encoder $\mathcal{E}$ to learn robust features against miss-detections. Specifically, we input $\hat{H}$ obtained from corrupted detection $X^C$ by \cref{eq:main-trajectory-prediction} into the unmasking head $\mathcal{P}_{M}$, as follows: $\hat{X}_{unmask} = \mathcal{P}_{M}(\hat{H}).$

\noindent \textbf{Denoising Task.} The goal of the denoising pretext task is to encourage the encoder to learn robust features that mitigate localization errors. Similarly, the denoising is formulated as follows: $\hat{X}_{denoise} = \mathcal{P}_{D}(\hat{H}).$

\noindent \textbf{Person Identity Reconstruction Task.} 
The person identity reconstruction pretext task aims to enhance the model's ability to effectively associate pedestrian instances across observation frames. To achieve this, the model is trained to reconstruct pedestrian identities from detections $X$ by predicting the person identity embeddings. The person identity reconstruction task is formulated as follows: $\hat{I} = \mathcal{P}_{R}(\hat{H}),$ 

where $\hat{I} \in \mathbb{R}^{KT_{obs} \times d_I}$ are the reconstructed person ID embeddings for past detections.

\noindent \textbf{Loss.}
The model is trained to minimize the losses for the four tasks mentioned above:
\begin{align}
\begin{split}
\mathcal{L} = \mathcal{L}_{F}& (\hat{Y}, Y) + \alpha \mathcal{L}_M (\hat{X}_{unmask}, X) \\ &+ \beta \mathcal{L}_D (\hat{X}_{denoise}, X) + \gamma \mathcal{L}_I (\hat{I}, I),
\end{split}
\end{align}
where $\mathcal{L}{_F}$, $\mathcal{L}_M$, $\mathcal{L}_D$, and $\mathcal{L}_I$ represent the trajectory forecasting\footnote{For the multi-future prediction, we employ winner-take-all strategy which only optimizes the best prediction with minimal average prediction error to the ground truth, following ~\cite{cheng2023forecastmae}.}, unmasking, denoising, and person identity reconstruction losses, respectively. We use the mean square error (MSE) loss between the targets and predictions for each task. The terms $\alpha$, $\beta$, and $\gamma$ are weight hyperparameters.

\subsection{Weakly-Supervised Fine-tuning Phase}
We fine-tune the model in a weakly-supervised manner by using future ground-truth detections $X_{T_{obs}+1:T_{obs}+T_{pred}}$, thereby reducing ID annotation costs in real-world data. To achieve this, we propose to employ two losses. The first loss is calculated between the predicted future position $\hat{y_t}$ and the closest ground-truth detection $d_t^c$ at each future timestep, summing over the $T_{pred}$ timesteps as follows:
\begin{equation}
d_t^c = \underset{m \in M} {\operatorname{argmin}} \| \hat{y}_t - d_t^m \|_2, 
\end{equation}
\begin{equation}
\mathcal{L}_{W} = \sum_{t=T_{obs} + 1}^{T_{obs} + T_{pred}}  \| \hat{y}_t - d_t^c \|_2.
\end{equation}

However, this formulation poses a challenge: the predicted position may sometimes deviate from the true trajectory due to its reliance on the closest detection, resulting in unintended oscillations. To address this issue, we introduce an acceleration regularization term, which discourages abrupt changes in acceleration. This promotes smoother transitions and reduces fluctuations in the predicted trajectories. This results in smoother, more realistic trajectory forecasting that aligns with natural motion dynamics. The acceleration regularization term is formulated as follows:
\begin{equation}
\mathcal{L}_{Reg} = \sum_{t=T_{obs} + 1}^{T_{obs} + T_{pred}-1}  \| \hat{y}_{t+1} - 2\hat{y}_t + \hat{y}_{t-1}\|_2.
\end{equation}
The overall loss function for fine-tuning is defined as:
\begin{equation}
\label{eq:weak-finetuning}
\mathcal{L}_{ft} = \mathcal{L}_{W} + \lambda  \mathcal{L}_{Reg},
\end{equation}
where the terms $\lambda$ is weight hyperparameter.

\begin{figure}[tb]
\centering
\resizebox{\columnwidth}{!}{
\begin{tabular}{cccc}
\begin{minipage}{0.25\textwidth}
    \begin{center}
      \includegraphics[clip, width=\hsize]{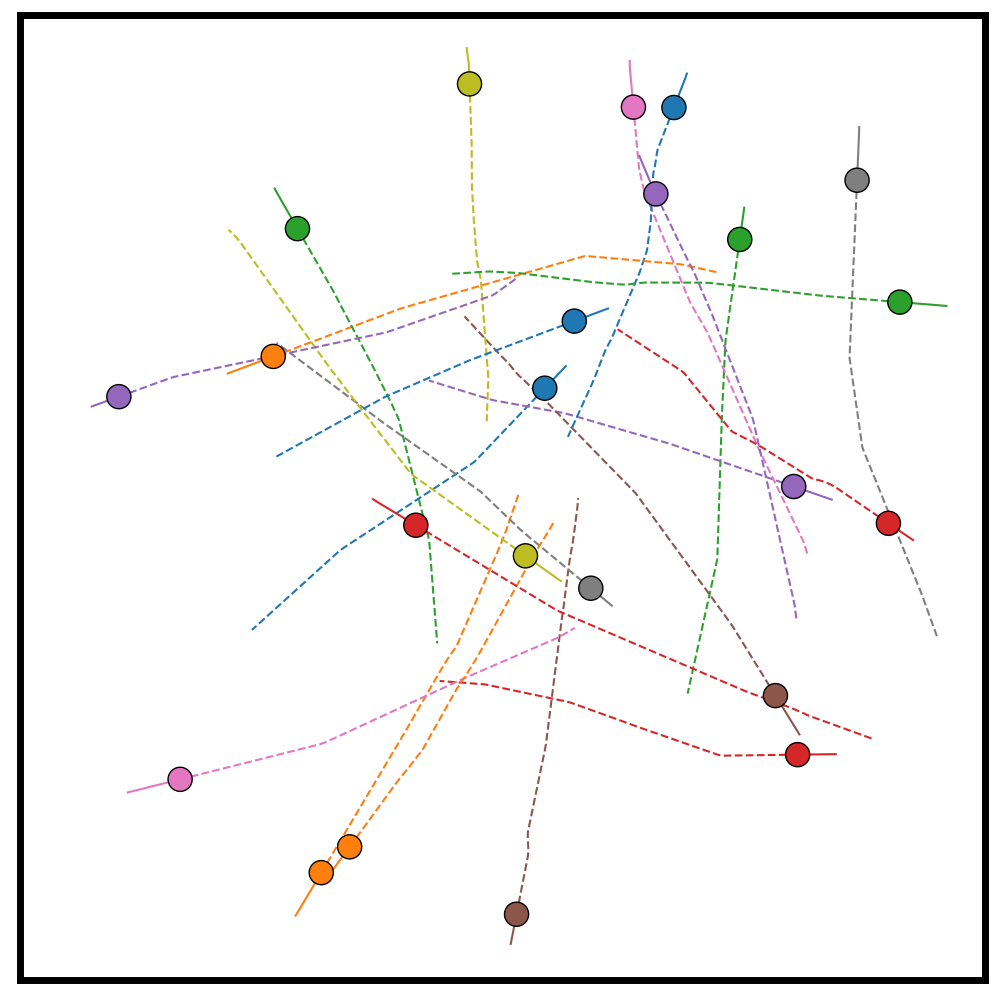}
    \end{center}
\end{minipage}
&
\begin{minipage}{0.25\textwidth}
    \begin{center}
      \includegraphics[clip, width=\hsize]{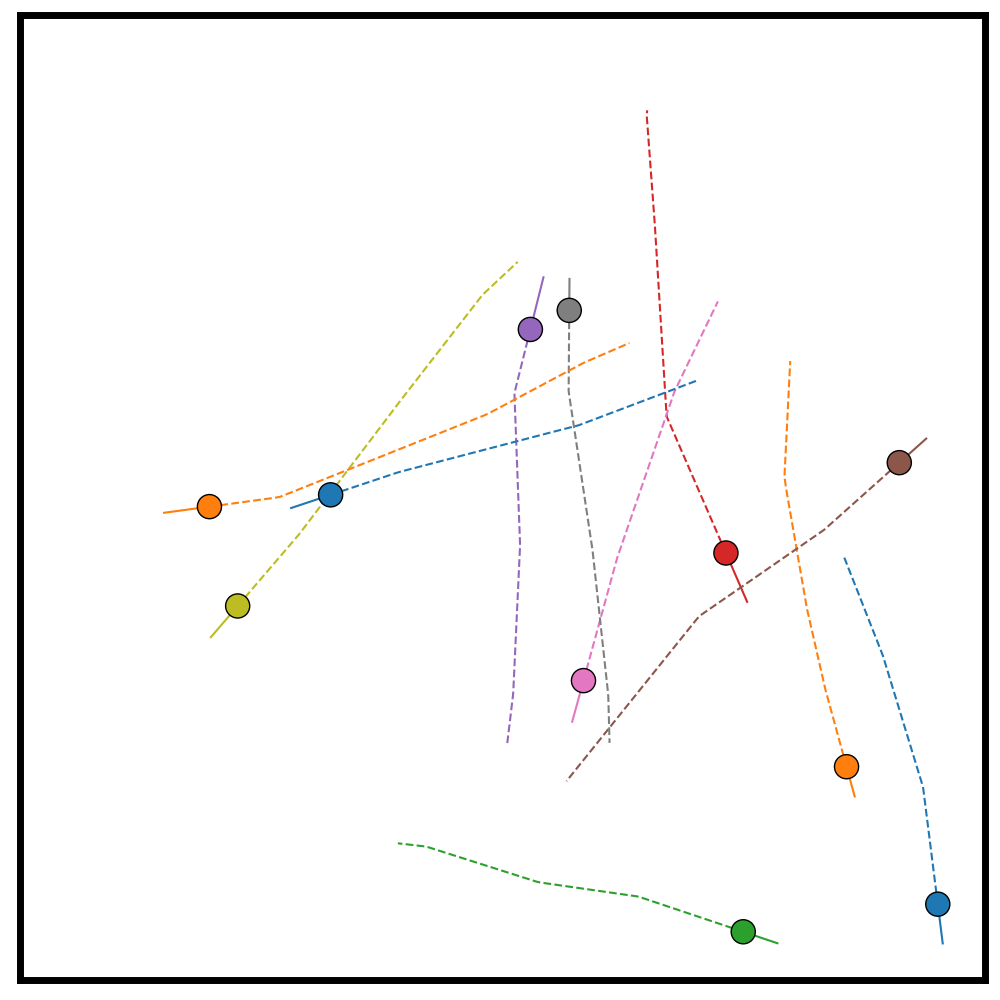}
    \end{center}
\end{minipage}
&
\begin{minipage}{0.25\textwidth}
    \begin{center}
      \includegraphics[clip, width=\hsize]{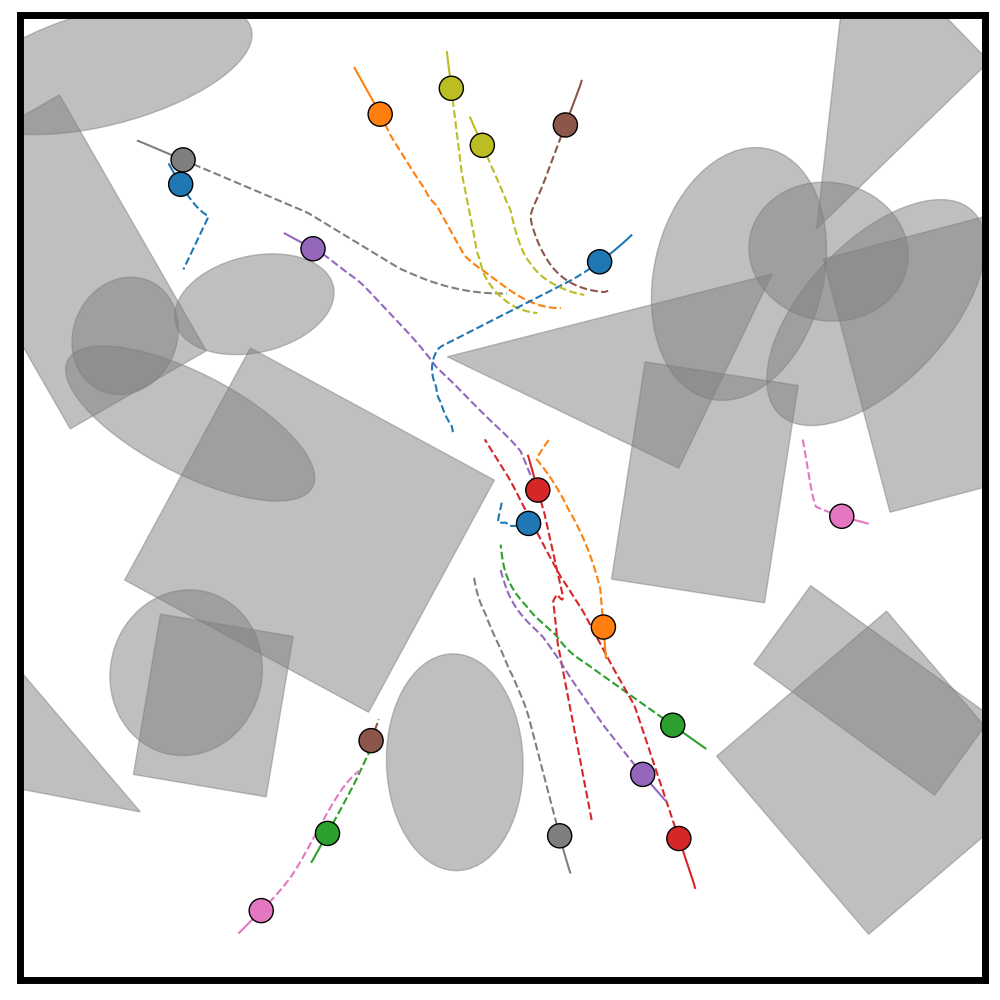}
    \end{center}
\end{minipage}
&
\begin{minipage}{0.25\textwidth}
    \begin{center}
      \includegraphics[clip, width=\hsize]{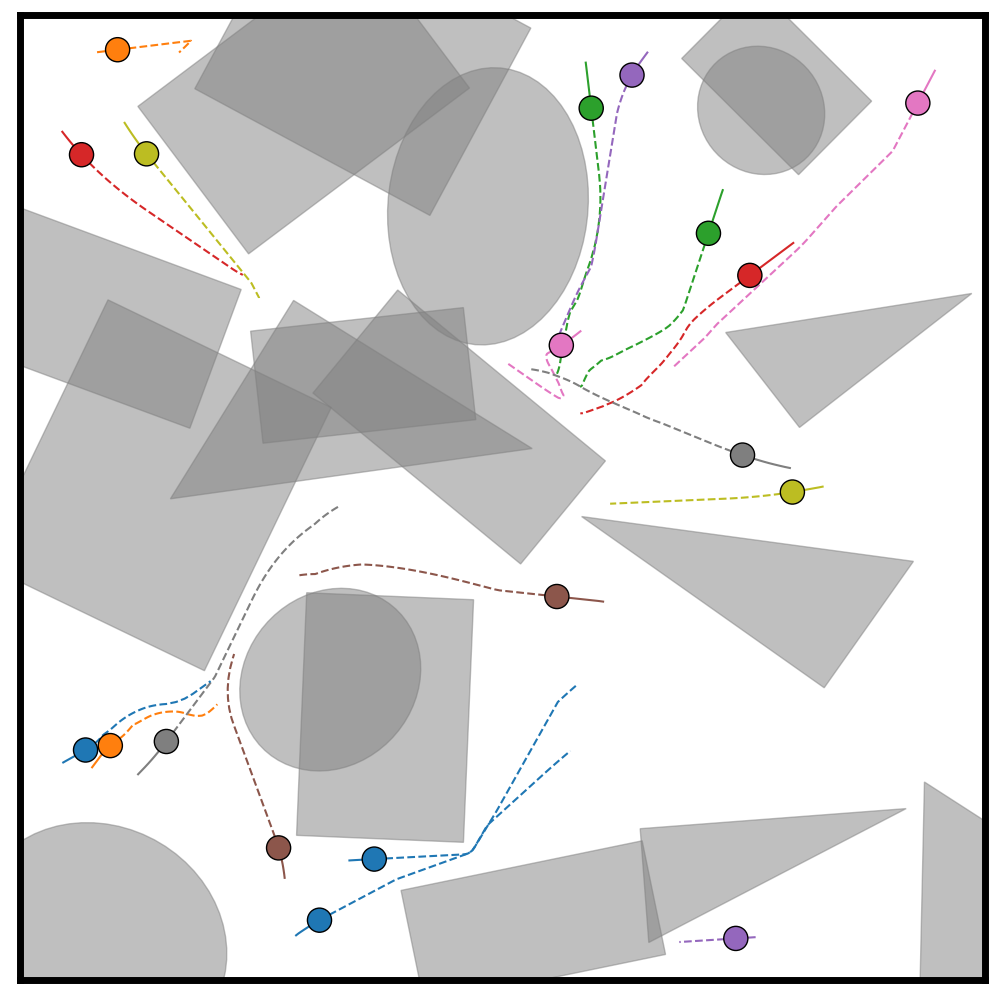}
    \end{center}
\end{minipage}
\end{tabular}}
\caption{Examples of synthetic trajectories generated for self-supervised pretraining.}
 \label{fig:example-synthetic-trajectories}
 \vspace{-1.em}
\end{figure}

\section{Experiments}
\begin{figure*}[tb]
\centering
\resizebox{\textwidth}{!}{
\begin{tabular}{cccc}
\begin{minipage}{0.25\textwidth}
    \begin{center}
        \includegraphics[clip, width=\hsize]{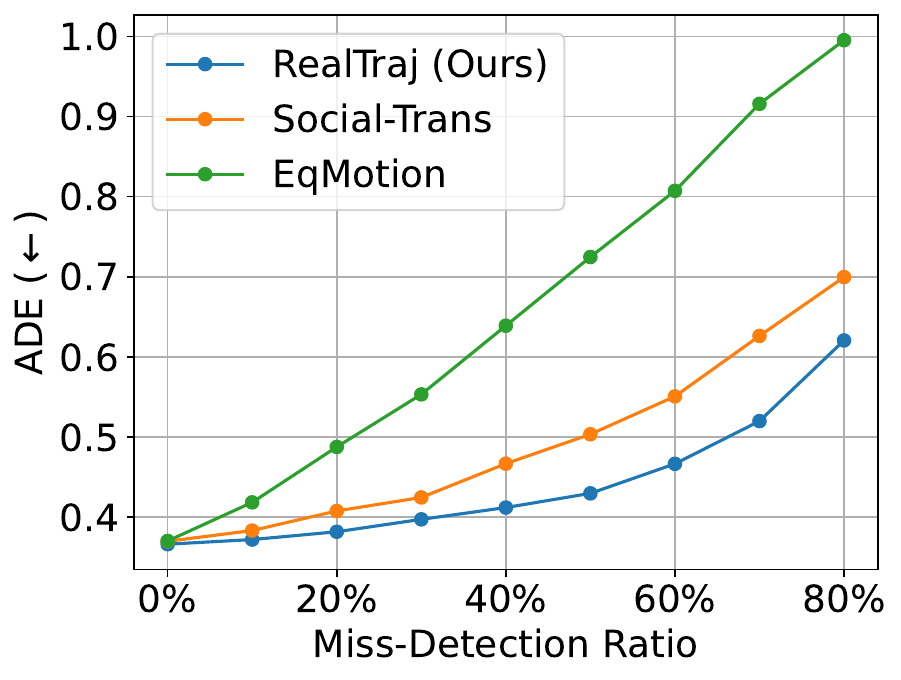} 
        (a) Miss-Detection
    \end{center}
\end{minipage}
&
\begin{minipage}{0.25\textwidth}
    \begin{center}
        \includegraphics[clip, width=\hsize]{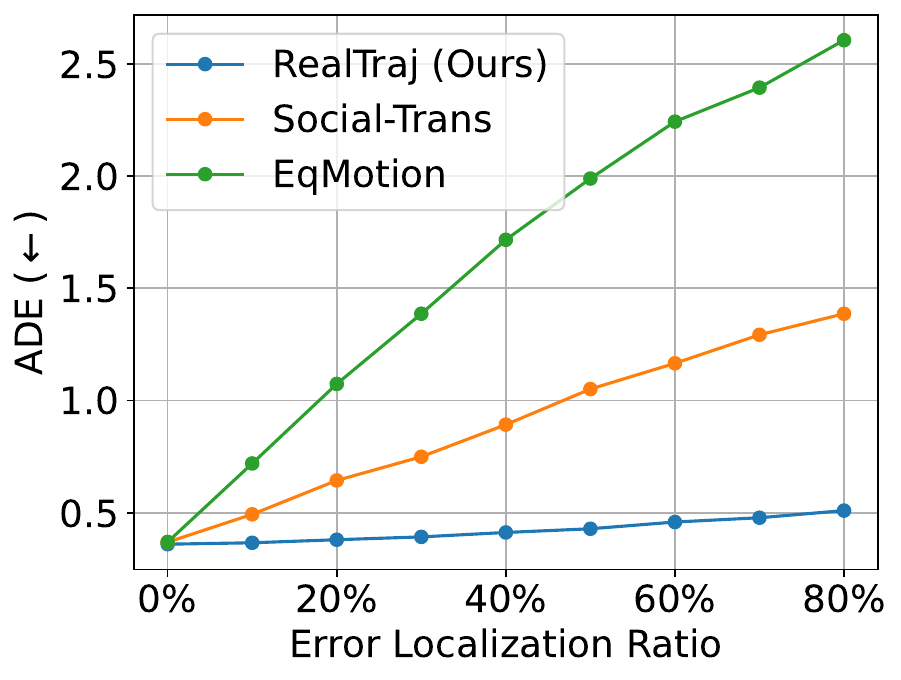} 
        (b) Error Localization
    \end{center}
\end{minipage}

&
\begin{minipage}{0.25\textwidth}
    \begin{center}
        \includegraphics[clip, width=\hsize]{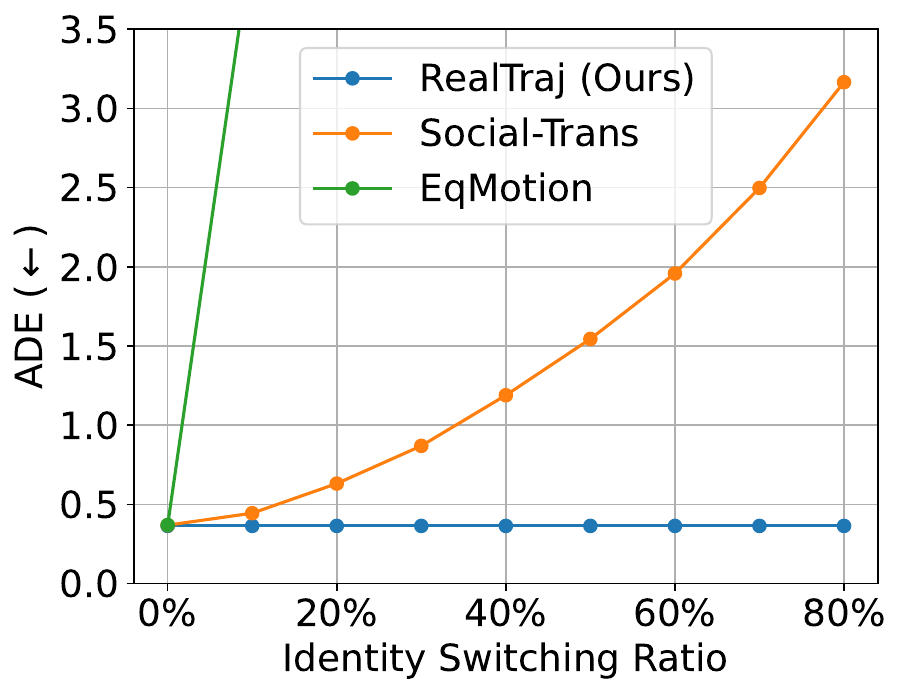} 
        (c) Identity Switch
    \end{center}
\end{minipage}

&
\begin{minipage}{0.25\textwidth}
    \begin{center}
        \includegraphics[clip, width=\hsize]{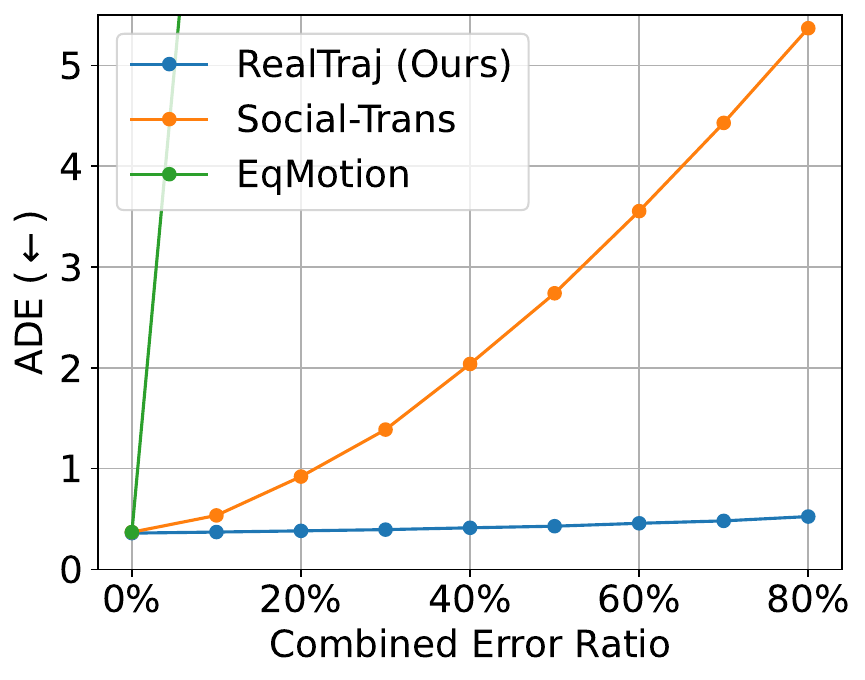} 
        (d) Combined Error
    \end{center}
\end{minipage}
\end{tabular}}
\caption{Comparison of model robustness against various error types on the JRDB dataset, including detection errors (miss-detections and localization errors), tracking errors (identity switches), and their combined impact.}
\label{fig:robustness}
\vspace{-1.em}
\end{figure*}

\subsection{Datasets}

\noindent \textbf{JackRabbot Dataset (JRDB)}: The JRDB dataset~\cite{martin2021jrdb}, recorded using a social robot, provides diverse real-world pedestrian trajectories in indoor and outdoor settings. We use the train-validation-test split from prior work~\cite{saadatnejad2024socialtransmotion}.

\noindent \textbf{Joint Track Auto Dataset (JTA)}: The JTA dataset~\cite{fabbri2018learning} is a large-scale synthetic dataset comprising $256$ training sequences, $128$ validation sequences, and $128$ test sequences set in urban scenarios, generated using a video game engine.

\noindent \textbf{ETH-UCY}: The ETH and UCY datasets~\cite{pellegrini2010eth,leal2014ucy} serve as established benchmarks for human trajectory forecasting, encompassing five scenes. We utilize the leave-one-out validation strategy as described in prior works~\cite{gupta2018socialgan,mangalam2021ynet}.

\noindent \textbf{Stanford Drone Dataset (SDD)}: The SDD dataset~\cite{robicquet2016sdd} features over $11,000$ unique pedestrians across $20$ scenes, captured from a bird's-eye view using a drone. Following the previous trajectory forecasting methods~\cite{mangalam2021ynet}, we use the standard setup and train-test split.

\noindent \textbf{TrajImpute}: The TrajImpute dataset~\cite{pranav2024missingbenchmark} simulates missing values within ETH-UCY, offering a framework for evaluating trajectory imputation and prediction models.

Following the standard practice, we predict $12$ future timesteps based on $9$ past timesteps for the JRDB and JTA datasets, and $12$ future timesteps based on $8$ past timesteps for the ETH-UCY, SDD, and TrajImpute datasets.

\subsection{Evaluation Metrics}
Average Displacement Error (ADE) and Final Displacement Error (FDE) are employed as evaluation metrics. ADE quantifies the average error over all predicted and ground-truth trajectory points, while FDE measures the displacement at the endpoints. Additionally, we report the minimum ADE and FDE (\text{minADE}$_N$, \text{minFDE}$_N$) over $N=20$ trajectories per pedestrian for multi-future prediction setting.

\subsection{Implementation Details}
We use the ORCA crowd simulator~\cite{berg2011ocra, rempeluo2023tracepace} to generate 2K synthetic trajectories, as shown in \cref{fig:example-synthetic-trajectories}. We train our model using the Adam optimizer~\cite{kingma2017adam}, starting with a learning rate of $1 \times 10^{-4}$, which is reduced by a factor of $0.1$ after completing $80\%$ of the total $200$ epochs for pretraining and $20$ epochs for fine-tuning. We set $\lambda=10$ for balancing the loss weight in \cref{eq:weak-finetuning}. See the supplementary material for more details (\eg, training hyperparameters).

\subsection{Robustness Evaluation}
\label{sec:robustness_protoc}
The robustness of the proposed method against perception errors (detection and tracking errors) is compared with that of the transformer-based (Social-Trans~\cite{saadatnejad2024socialtransmotion}) and graph-based (EqMotion~\cite{xu2023eqmotion}) models, which are the well-known state-of-the-art methods across multiple benchmarks (\textit{c.f.}, \cref{tab:comparison-with-sotas}). Although Social-Trans is designed to process multiple modalities, we used only trajectory data as input in all experiments to ensure fair comparisons. We synthetically generated two types of detection errors—miss-detections and localization errors—along with one type of tracking error, specifically identity switches, and combined these errors for evaluation. Miss-detections were simulated by setting past coordinates to zero for our model and linearly interpolated for comparison models, while localization errors were introduced by adding Gaussian noise to the inputs. Identity switches were simulated by swapping identities with nearby pedestrians within a 5-meter radius. 
Both detection and tracking errors were introduced for combined errors. 

{\cref{fig:robustness} (a, b)} shows that the performance of comparison methods significantly degrades when detection errors are introduced. In contrast, the proposed method exhibits relatively minor performance degradation, demonstrating robustness against detection errors. {\cref{fig:robustness} (c)} presents a comparison of robustness against tracking errors. While the comparison methods show high sensitivity to tracking errors, our proposed method remains unaffected due to its network architecture, Det2TrajFormer, which forecasts future trajectories directly from detection inputs without relying on tracking information. Lastly, in {\cref{fig:robustness} (d)}, we compare robustness against combined detection and tracking errors, where RealTraj consistently outperforms comparison methods by a significant margin across all error ratios.

\begin{table}[tb]
 \caption{Comparison with the current state-of-the-art method on the Hard-Impute subset of TrajImpute. We report the minADE$_{20}$ metric. The unit for minADE$_{20}$ is meters. The best results are \textbf{bolded}, and the second-best results are \underline{underlined}. Note that our model takes detections as inputs, while other approaches take the trajectory as inputs.}
\begin{center}
\vspace{-0.5em}
\resizebox{\columnwidth}{!}{
\begin{tabular}{lccccccc}
\toprule
Method & ETH & HOTEL & UNV & ZARA1 & ZARA2 & Avg. \\
\toprule
GPGraph~\cite{bae2022gpgraph} & 0.92 & 1.89 & 0.53 & 0.58 & 0.36 & 0.86 \\
GraphTern\cite{bae2023graphtern} & 0.78 & 1.68 & 0.50 & 0.96 & 0.37 & 0.86 \\
LBEBM-ET~\cite{bae2023eigentrajectory} & 0.85 & 3.31 & 0.64 & 0.37 & 0.27 & 1.09 \\
SGCN-ET~\cite{bae2023eigentrajectory} & 1.07 & 3.21 & 0.77 & 0.61 & 0.41 & 1.21 \\
TUTR~\cite{shi2023tutr} & 1.12 & 3.36 & 0.59 & 0.50 & \underline{0.33} & 1.18 \\ 
EqMotion~\cite{xu2023eqmotion} & \textbf{0.47} & \underline{0.72} & \underline{0.39} & \textbf{0.28} & 0.37 & \underline{0.45} \\
\hhline{|-|-|-|-|-|-|-|}%
\rowcolor{Gray}RealTraj (Ours) & \underline{0.48} & \textbf{0.20} & \textbf{0.37} & \textbf{0.28} & \textbf{0.21} & \textbf{0.30} \\ 
\bottomrule
\end{tabular}}
\end{center}
 \label{tab:trajimpute}
 \vspace{-1.5em}
\end{table}

Additionally, we evaluate RealTraj's robustness against missing data (\ie, miss-detection) using the TrajImpute~\cite{pranav2024missingbenchmark}. Since the combination of the imputation method SAITS~\cite{du2023saits} and EqMotion achieved the best performance in the TrajImpute benchmark, we adopt SAITS as the imputation method in this evaluation. The results, presented in~\cref{tab:trajimpute}, show that RealTraj outperforms all comparison methods across most subsets. This includes significant improvements on the challenging HOTEL subset, where the imputation is especially difficult~\cite{pranav2024missingbenchmark}. Additional result under the momentary trajectory prediction and observation length shift settings is provided in the supplementary materials. These findings confirm that the proposed model significantly enhances robustness against perception errors, aligning with our goal of addressing the first robustness limitation described in~\cref{sec:intro}.

\begin{table*}[tb]
\caption{Comparisons on JRDB, JTA, ETH-UCY, and SDD under few-shot evaluation setting. The ADE metric is reported with different percentages of labeled real data during fine-tuning. The S and WS in the Setting column indicate that the models are trained in a fully-supervised and weakly-supervised manner, respectively. The T and D in the Input column denote that the models take trajectories and detections as inputs, respectively. The column (w/ Syn.) indicates whether the models are pretrained on the synthetic trajectory data. The unit for ADE is meters for JRDB, JTA, and ETH-UCY, and pixels for SDD.
The best results are \textbf{bolded}, and the second-best results are \underline{underlined}.}
\begin{center}
\vspace{-1.em}
\resizebox{\textwidth}{!}{
\begin{tabular}{cc}
\resizebox{\linewidth}{!}{
\begin{tabular}{c}
\large{(a) JRDB}
\\
\begin{tabular}{cclc|ccccc}
\toprule
Setting  & Input & Method & w/ Syn. & 0.1$\%$ & 0.5$\%$ & 1$\%$ & 2$\%$ & 5$\%$   \\
 \toprule
  \multirow{6}{*}{S} & \multirow{4}{*}{T} & \multirow{2}{*}{EqMotion~\cite{xu2023eqmotion}}  & \xmark  & \underline{0.45} & 0.48 & 0.43 &  0.60  & 0.43 \\ 
&  & & \checkmark & 0.46 & \underline{0.45} & \underline{0.42}  & 0.41 & \underline{0.41} \\  \hhline{|~|~|-|-|-|-|-|-|-|}%
 &  & \multirow{2}{*}{Social-Trans~\cite{saadatnejad2024socialtransmotion}} & \xmark & 0.68  & 0.48 & 0.47  & 0.43 & 0.43 \\ 
 & & & \checkmark & 0.61  & \underline{0.45} & 0.43  & 0.42 & \underline{0.41} \\ 
 \hhline{|~|-|-|-|-|-|-|-|-|}%
\rowcolor{Gray} &  &   & \xmark  & 0.92 &0.88 & 0.88 & 0.87 & 0.86 \\
  & \multirow{-2}{*}{D} &  \multirow{-2}{*}{RealTraj (Ours)} & \checkmark  & \textbf{0.43} & \textbf{0.41} & \textbf{ 0.40} & \textbf{0.38} & \textbf{0.38} \\ \hline
 &  &   & \xmark & 1.11 & 0.89 & 0.88  & 0.87 & 0.86  \\
  \multirow{-2}{*}{WS} & \multirow{-2}{*}{D} & \multirow{-2}{*}{RealTraj (Ours)}  & \checkmark & \textbf{0.43}& \textbf{0.41} & \textbf{0.40}  & \underline{0.39} & \textbf{0.38}  \\
\bottomrule
\end{tabular}
\end{tabular}}
& 
\resizebox{\linewidth}{!}{
\begin{tabular}{c}
\large{(b) JTA}
\\
\begin{tabular}{cclc|ccccc}
\toprule
Setting  & Input & Method & w/ Syn. & 0.1$\%$ & 0.5$\%$ & 1$\%$ & 2$\%$ & 5$\%$   \\
 \toprule
  \multirow{6}{*}{S} & \multirow{4}{*}{T} &  \multirow{2}{*}{EqMotion~\cite{xu2023eqmotion}}  & \xmark  & 1.76 & 1.68 &  1.55 & 1.55 & 1.55 \\
& & & \checkmark & \underline{1.75} & \underline{1.55} & 1.53  & 1.46 & 1.43 \\ \hhline{|~|~|-|-|-|-|-|-|-|}%
 &  & \multirow{2}{*}{Social-Trans~\cite{saadatnejad2024socialtransmotion}}  & \xmark & 4.80 & 3.33 & 2.91 & 2.78  & 2.78 \\ 
 & & & \checkmark & 2.48 & 1.63 & \underline{1.49} & \underline{1.39} & 1.29 \\  \hhline{|~|-|-|-|-|-|-|-|-|}
\rowcolor{Gray}&  &  & \xmark  & 4.83 & 4.74 & 4.67 & 1.43 & 1.37 \\
  & \multirow{-2}{*}{D} & \multirow{-2}{*}{RealTraj (Ours)} & \checkmark  &  \textbf{1.38} & \textbf{1.25} & \textbf{1.22} & \textbf{1.20} & \textbf{1.07}   \\ \hline
& & & \xmark & 4.98 & 4.93 & 4.91 & 4.88 & 4.86 \\
  \multirow{-2}{*}{WS} &  \multirow{-2}{*}{D} &  \multirow{-2}{*}{RealTraj (Ours)}  & \checkmark &2.13 &  1.98 & 1.68 & 1.46 & \underline{1.23}   \\
\bottomrule
\end{tabular}
\end{tabular}}
\vspace{1.em}
\\
\resizebox{\linewidth}{!}{
\begin{tabular}{c}
\large{(c) ETH-UCY}
\\
\begin{tabular}{cclc|ccccc}
\toprule
Setting  & Input & Method & w/ Syn. & 0.1$\%$ & 0.5$\%$ & 1$\%$ & 2$\%$ & 5$\%$   \\
 \toprule
  \multirow{6}{*}{S} & \multirow{4}{*}{T} &  \multirow{2}{*}{EqMotion~\cite{xu2023eqmotion}}   & \xmark  & 0.70 &\underline{0.56}  &  \underline{0.53} &  \underline{0.53} &  0.53\\
&  & & \checkmark & 0.81 & 0.60 & 0.60 & 0.61 & 0.56 \\  \hhline{|~|~|-|-|-|-|-|-|-|}%
 &  & \multirow{2}{*}{Social-Trans~\cite{saadatnejad2024socialtransmotion}} & \xmark & 1.05  & 0.62 & 0.55 & \textbf{0.52}  & 0.53 \\ 
 & & & \checkmark & \underline{0.57} & \underline{0.56} & \textbf{0.52} & \textbf{0.52} & \textbf{0.50} \\ \hhline{|~|-|-|-|-|-|-|-|-|}%
\rowcolor{Gray}&  & & \xmark & 0.81 & 0.72 & 0.60 & 0.61  & 0.57 \\
  & \multirow{-2}{*}{D} &  \multirow{-2}{*}{RealTraj (Ours)}  & \checkmark  &  \textbf{0.56}  &\textbf{0.55}  & \underline{0.53}& \textbf{0.52} & \underline{0.52}   \\ \hline
& & & \xmark & 1.37 & 1.13 & 0.81 & 0.67 & 0.57 \\
 \multirow{-2}{*}{WS} &  \multirow{-2}{*}{D} &  \multirow{-2}{*}{RealTraj (Ours)}  & \checkmark & \textbf{0.56} & \underline{0.56} & 0.56 & \underline{0.53} & 0.53    \\
\bottomrule
\end{tabular}
\end{tabular}}

&
\resizebox{\linewidth}{!}{
\begin{tabular}{c}
\large{(d) SDD}
\\
\begin{tabular}{cclc|ccccc}
\toprule
Setting  & Input & Method & w/ Syn. & 0.1$\%$ & 0.5$\%$ & 1$\%$ & 2$\%$ & 5$\%$   \\
 \toprule
  \multirow{6}{*}{S} & \multirow{4}{*}{T} & \multirow{2}{*}{EqMotion~\cite{xu2023eqmotion}}    & \xmark  & 24.7 & \underline{20.0} & \underline{18.5} & 19.0 &  17.6  \\ 
& & & \checkmark & 25.8 & 20.9 &  19.3 & 19.2 & 19.8 \\ \hhline{|~|~|-|-|-|-|-|-|-|}%
 & & \multirow{2}{*}{Social-Trans~\cite{saadatnejad2024socialtransmotion}} & \xmark & 45.0 & 32.7 & 20.5 & 21.8  & 19.0  \\ 
 &  &  & \checkmark & 31.5 & \textbf{18.1} & \textbf{17.9} & \textbf{17.4} & \underline{17.3} \\ \hhline{|~|-|-|-|-|-|-|-|-|}%
\rowcolor{Gray}&  & & \xmark  & 67.4 & 41.8 & 38.3 & 36.3 & 18.7  \\
  & \multirow{-2}{*}{D} & \multirow{-2}{*}{RealTraj (Ours)} & \checkmark  & \textbf{22.3} & 20.2 & 19.2 & \underline{18.5} &\textbf{ 17.0}  \\ \hline
& & & \xmark & 70.6 & 54.4 & 41.2 & 38.7 & 19.5 \\
 \multirow{-2}{*}{WS} &  \multirow{-2}{*}{D} &  \multirow{-2}{*}{RealTraj (Ours)} & \checkmark &\underline{23.5}  & 20.3 & 19.7 & 19.0 &17.9 \\
\bottomrule
\end{tabular}
\end{tabular}}
\end{tabular}}
\end{center}
 \label{tab:comparison-few-shot-setting}
 \vspace{-1.em}
\end{table*}

\begin{table*}[tb]
 \caption{Comparisons with state-of-the-art methods on the JTA, JRDB, ETH-UCY, and SDD under fully-supervised setting. The S and WS in the Setting column indicate that the models are trained in a fully-supervised and weakly-supervised manner, respectively. The T and D in the Input column denote that the models take trajectories and detections as inputs, respectively. The unit for ADE and FDE is meters for JRDB, JTA, and ETH-UCY, and pixels for SDD. The best results are \textbf{bolded}, and the second-best results are \underline{underlined}. \textdagger: We employ Social-Transmotion with trajectory inputs for a fair comparison. For the multi-future prediction setting, we extend it to generate multiple future trajectories using the same approach as ours.}
\begin{center}
\vspace{-0.5em}
\resizebox{\textwidth}{!}{
\begin{tabular}{cclccccccccccccc}
\toprule
\multirow{2.5}{*}{Setting}  & \multirow{2.5}{*}{Input}  &  \multirow{2.5}{*}{Method}   &\multicolumn{2}{c}{JRDB}   &\multicolumn{2}{c}{JTA} &\multicolumn{4}{c}{ETH-UCY} &\multicolumn{4}{c}{SDD} \\
 \cmidrule(l{2pt}r{3pt}){4-5} \cmidrule(l{2pt}r{3pt}){6-7} \cmidrule(l{2pt}r{3pt}){8-11} \cmidrule(l{2pt}r{3pt}){12-15}
& & & ADE & FDE & ADE & FDE & ADE & FDE & minADE$_{20}$ & minFDE$_{20}$ & ADE & FDE& minADE$_{20}$ & minFDE$_{20}$\\
\toprule
 \multirow{13}{*}{S} 
& \multirow{12}{*}{T} & Social-LSTM~\cite{alahi206sociallstm} & 0.47 & 0.95 & 1.21 & 2.54 & 0.72 & 1.54 & - &- & - & -   & -  & -\\
& & Social-GAN-ind~\cite{gupta2018socialgan} & 0.50 & 0.99 & 1.66 & 3.76 & 0.74 & 1.54 & 0.58 & 1.18 & - & - & 27.2 & 41.4 \\

& & Trajectron++~\cite{salzman2020trajectronplusplus} & 0.40 & 0.78 & 1.18 & 2.53 & 0.53 & 1.11 & \textbf{0.21} & 0.41& - & - & - & -\\
& & Transformer~\cite{giuliari2021transformer} &  0.56 & 1.10 & 1.56 & 3.54 & 0.54 & 1.17 & 0.31 & 0.55 & - & -  & - & - \\
& & Directional-LSTM~\cite{kothari2022human} & 0.45 & 0.87 & 1.37 & 3.06 & - & - & - & - & - & - & - & - \\

& & Autobots~\cite{girgis2022autobot} & 0.39 & 0.80  & 1.20 & 2.70 & 0.52 & \underline{1.05} & - &- & - & - & - & - \\
& & TUTR~\cite{shi2023tutr} & - & - & - & - & 0.53 & 1.12 & \textbf{0.21} & 0.36 & 17.4 & 35.0 & 7.76 & 12.7 \\
& & EigenTrajectory~\cite{bae2023eigentrajectory} & - & - & - & - & \underline{0.51} & 1.11 & \underline{0.22} & 0.37 & 20.7 & 41.9 & 8.12 & 13.1 \\
& & FlowChain~\cite{Maeda_2023_ICCV} & - & - & - & - & - & - & 0.29 & 0.52 & - & - & 9.93 & 17.17 & \\
& & EqMotion~\cite{xu2023eqmotion} & 0.42 & 0.78 & 1.13 & 2.39  & \textbf{0.49} & \textbf{1.03} & \textbf{0.21} & \underline{0.35} & 16.8 & 33.7 & 8.45 & 14.1 \\

  & & Social-Trans\textsuperscript{\textdagger}~\cite{saadatnejad2024socialtransmotion} & 0.40 & 0.77 & \underline{0.99} & \underline{1.98} & \underline{0.51} & \textbf{1.03} & \textbf{0.21} & 0.41 & 18.0 & 38.7 & \underline{7.21} & 14.3 \\
 & & MGF~\cite{Chen2024MGF} & - & - & - & - & - & - & \textbf{0.21} & \textbf{0.34} & - & - & 7.74 &  \textbf{12.1}
 \\ \hhline{|~|-|-|-|-|-|-|-|-|-|-|-|-|-|-|}%
 \rowcolor{Gray} & D & RealTraj (Ours)  & \textbf{0.35} & \textbf{0.67}  & \textbf{0.92} & \textbf{1.84} & \underline{0.51} & \textbf{1.03} & 0.23 & 0.42 & \textbf{16.0} & \textbf{31.8} & \textbf{7.18} & \underline{12.4} \\ \hhline{|-|-|-|-|-|-|-|-|-|-|-|-|-|-|-|}%
 \rowcolor{Gray}  WS & D & RealTraj (Ours)  &  \underline{0.36} & \underline{0.71}  & 1.04 & 2.11 & 0.52 & 1.07 &  0.26 & 0.43 &\underline{16.0} & \underline{32.0} & 8.21 & 13.8 \\
\bottomrule
    \end{tabular}}
\end{center}
 \label{tab:comparison-with-sotas}
 \vspace{-1.em}
\end{table*}

\subsection{Few-shot Evaluation}
Next, we demonstrate the effectiveness of RealTraj with limited real-world labeled data. We randomly selected subsets of $0.1\%$, $0.5\%$, $1\%$, $2\%$, and $5\%$ of the datasets, fine-tuned the model on each subset, and evaluated it on the test set on JRDB, JTA, ETH-UCY, and SDD. In this experiment, since no prior work focuses on trajectory prediction with a few-shot setting, we compare our model against the existing approaches, EqMotion~\cite{xu2023eqmotion} and Social-Trans~\cite{saadatnejad2024socialtransmotion}. We prepare the comparison models that are pretrained on the synthetic trajectory data, the same as ours, for a fair comparison. 

As shown in~\cref{tab:comparison-few-shot-setting}, our method consistently outperforms these prior methods on JRDB. In the most challenging scenario with only $0.1\%$ of annotations, weakly-supervised RealTraj (last row) achieves improvements of $4.4\%$, $20\%$, and $4.9\%$ on JRDB, ETH-UCY, and SDD, respectively, compared to the best-performing previous supervised model. In other data settings, RealTraj achieves results comparable to supervised methods despite the detection inputs and requiring only weak detection annotations during fine-tuning. This demonstrates that our model overcomes the second limitation of real-world data collection cost mentioned in  ~\cref{sec:intro}.
Furthermore, the weakly-supervised RealTraj (last row) achieved the performance on par with the fully-supervised RealTraj on JRDB and SDD datasets, verifying that our approach reduces the person ID annotation cost mentioned in ~\cref{sec:intro}.

\subsection{Comparison with State-of-the-art}
\label{sec:100_percent_protoc}
We further compare RealTraj with a diverse range of existing approaches under the standard 100\% labeled setting, as summarized in~\cref{tab:comparison-with-sotas}. Although achieving state-of-the-art performance in a fully supervised setting with clean trajectory inputs is not the primary focus of our paper, the results show that our weakly supervised framework performs comparably to fully supervised approaches using only detection inputs on the various benchmarks. Additionally, our fully supervised approach surpasses all other methods on the JRDB, JTA, and SDD datasets in terms of ADE and remains competitive on the ETH-UCY dataset.

\subsection{Ablation Study}

\begin{table}[tb]
 \caption{Ablation study of the proposed pretext tasks on the JRDB dataset. 'F', 'P', 'U', and 'D' represent forecasting, person ID reconstruction, unmasking, and denoising pretext tasks, respectively. The first row denotes the results without pretraining phase. Results are reported for both complete input and corrupted input scenarios with detection errors. The reported metric is ADE.}
  
\begin{center}
\vspace{-0.5em}
\resizebox{\columnwidth}{!}{
\begin{tabular}{ccccccc}
\toprule
Main Task  & \multicolumn{3}{c}{Pretext Tasks} &  \multirow{2.5}{*}{Complete} &\multicolumn{2}{c}{Detection Errors}   \\
\cmidrule(l{2pt}r{3pt}){1-1} \cmidrule(l{2pt}r{3pt}){2-4} \cmidrule(l{2pt}r{3pt}){6-7}
 F& P & U & D &   & Miss-Detection & Error Localization \\
 \midrule
 -  &- & - & -  & 0.78 & 0.91 & 0.84 \\
\checkmark  &- & - & - & 0.42 & 0.53 & 0.70 \\
-  &  \checkmark  & \checkmark & \checkmark & 0.39 &  0.47 & 0.51  \\
\checkmark  &  - & \checkmark & \checkmark & 0.42 & 0.52 & 0.62 \\
\checkmark & \checkmark & - & - & 0.41  & 0.51 & 0.54 \\
\checkmark  & \checkmark  & - & \checkmark & 0.38 & 0.47 & 0.51 \\
\checkmark  & \checkmark  & \checkmark & - & 0.38  & 0.44 & 0.45 \\
\rowcolor{Gray} \checkmark  & \checkmark  & \checkmark & \checkmark & \textbf{0.36}  & \textbf{0.42} & \textbf{0.43} \\ 
\bottomrule
\end{tabular}}
\vspace{-1.5em}
\end{center}
 \label{tab:ablation-pretext-tasks}
\end{table}

\noindent \textbf{Effectiveness of Pretext Tasks.} We first examine the effect of incorporating our proposed pretext tasks during pretraining on the JRDB, as shown in \cref{tab:ablation-pretext-tasks}. In addition to using complete inputs, we conducted ablation with inputs containing detection errors to demonstrate the contribution of each pretext task to robustness. Our results show that all pretext tasks enhance trajectory forecasting performance. First, the performance significantly degrades when removing the pretraining stage even with the complete input, verifying the effectiveness of pretraining for enabling the trajectory prediction with detections. Moreover, the person identity reconstruction task improves ADE by $7.9\%$ (from $0.39$ to $0.36$). Both the unmasking and denoising tasks also contribute to performance gains, each enhancing ADE by $5.3\%$ (from $0.38$ to $0.36$). Joint training with all three pretext tasks and the main forecasting task improves ADE by $14\%$ (from $0.42$ to $0.36$). 

The pretext tasks also contribute to robustness against detection errors. Specifically, the unmasking pretext task improves ADE by $11\%$ when using inputs with miss-detection errors, while the denoising pretext task enhances ADE by $4.4\%$ when using inputs with localization errors. Combining the unmasking and denoising tasks results in further improvements, boosting ADE by $18\%$ (from $0.51$ to $0.42$)  for inputs with miss-detections and by $20\%$ (from $0.51$ to $0.43$) for inputs with localization errors. Notably, pretraining is conducted using synthetic trajectories, achieving these results without additional data collection.

\begin{table}[tb]
 \caption{Ablation study of acceleration regularization weight $\lambda$ on the JTA to examine the impact of the acceleration regularization on overall performance. We selected $\lambda = 10$ as the default setting. }
\begin{center}
\vspace{-0.5em}
\resizebox{0.56\columnwidth}{!}{
\begin{tabular}{ccccc}
\toprule
\multirow{2}{*}{Metric}  & \multicolumn{4}{c}{$\lambda$}   \\
 \cmidrule(l{2pt}r{3pt}){2-5}
 & 0 &  1 & 10  & 100 \\
 \midrule
 ADE  & 1.16 & 1.09 & \textbf{1.04} & 3.67 \\
 FDE  & 2.32 & 2.18 &  \textbf{2.10} & 5.78 \\
\bottomrule
\end{tabular}
}
\vspace{-1.em}
\end{center}
 \label{tab:ablation-acceleration-regulalization}
\end{table}

\noindent \textbf{Effectiveness Acceleration Regularization.} Hyperparameter acceleration regularization $\lambda$  is used to balance the two terms in~\cref{eq:weak-finetuning}.
We conducted ablations with different values of acceleration regularization $\lambda$ on JTA in~\cref{tab:ablation-acceleration-regulalization}. The results indicate that setting $\lambda = 10$ provides the best performance for both ADE and FDE metrics, yielding a $10\%$ improvement in ADE (from $1.16$ to $1.04$) compared to performance without acceleration regularization ($\lambda = 0$).

\begin{figure}[tb]
\centering
\resizebox{\columnwidth}{!}{
\begin{tabular}{cc}
\begin{minipage}{0.5\columnwidth}
    \centering
    \includegraphics[clip, width=\hsize]{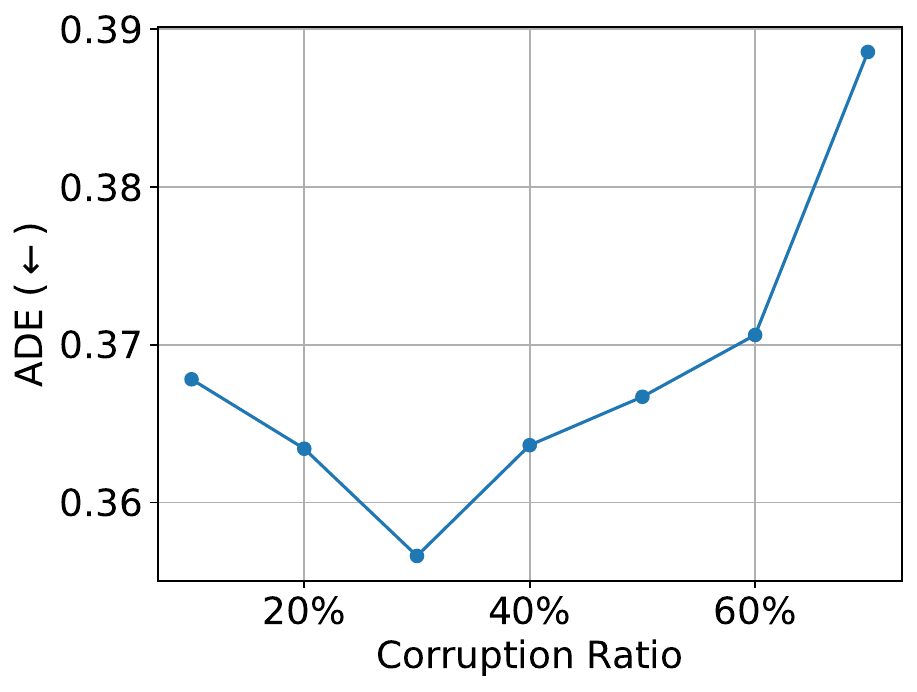} 
    {(a) Effect of corruption ratio during training.}
\end{minipage}
&
\begin{minipage}{0.5\columnwidth}
    \centering
    \includegraphics[clip, width=\hsize]{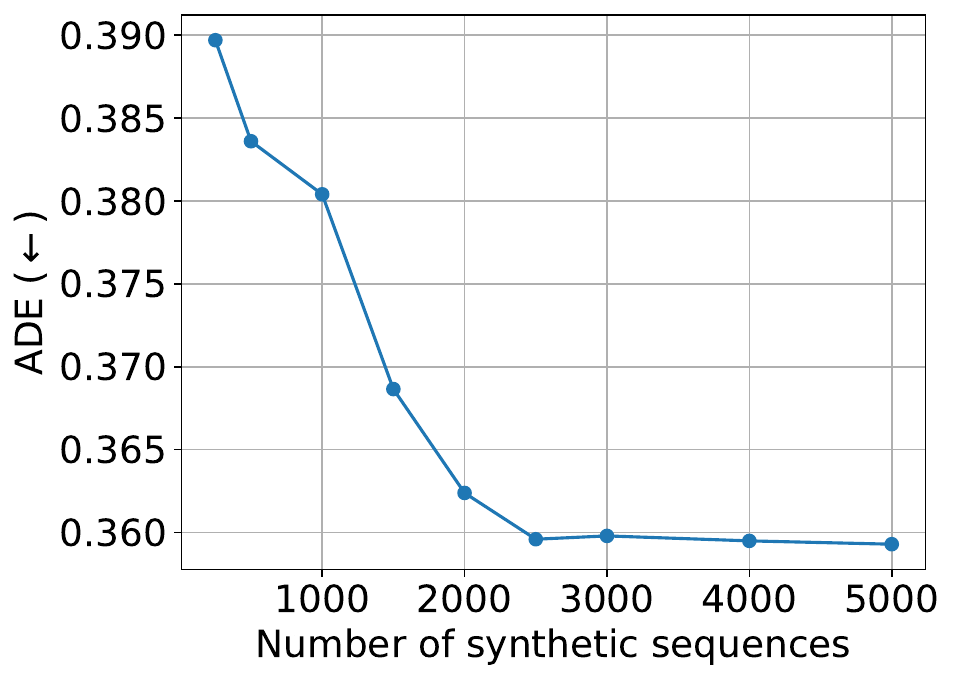} 
    {(b) Effect of the number of synthetic sequences.}
\end{minipage}
\end{tabular}
}

\caption{Effect of corruption ratio during training and number of synthetic trajectories.}
\label{fig:effect-of-corrupt-ratio-and-num-of-synth-traj}
\vspace{-1.em}
\end{figure}

\noindent \textbf{Effect of Corruption Ratio.}
We explore the effect of different corruption (adding noise and masking) ratios during pretraining in {\cref{fig:effect-of-corrupt-ratio-and-num-of-synth-traj} (a)}. Setting the corruption ratios either too low or too high results in suboptimal performance, as the pretext tasks become either too easy or too challenging, limiting their effectiveness for model learning. A moderate corruption ratio yielded the best results.

\noindent \textbf{Effect of the Number of Synthetic Sequences.}
{\cref{fig:effect-of-corrupt-ratio-and-num-of-synth-traj} (b)} presents the ADE transition when the number of pretraining samples is varied. Our framework shows improvement as the amount of data increases.

\begin{figure}[tb]
\centering
\resizebox{\columnwidth}{!}{
\begin{tabular}{cc}
\begin{minipage}{0.5\columnwidth}
    \begin{center}
        \includegraphics[clip, width=\hsize]{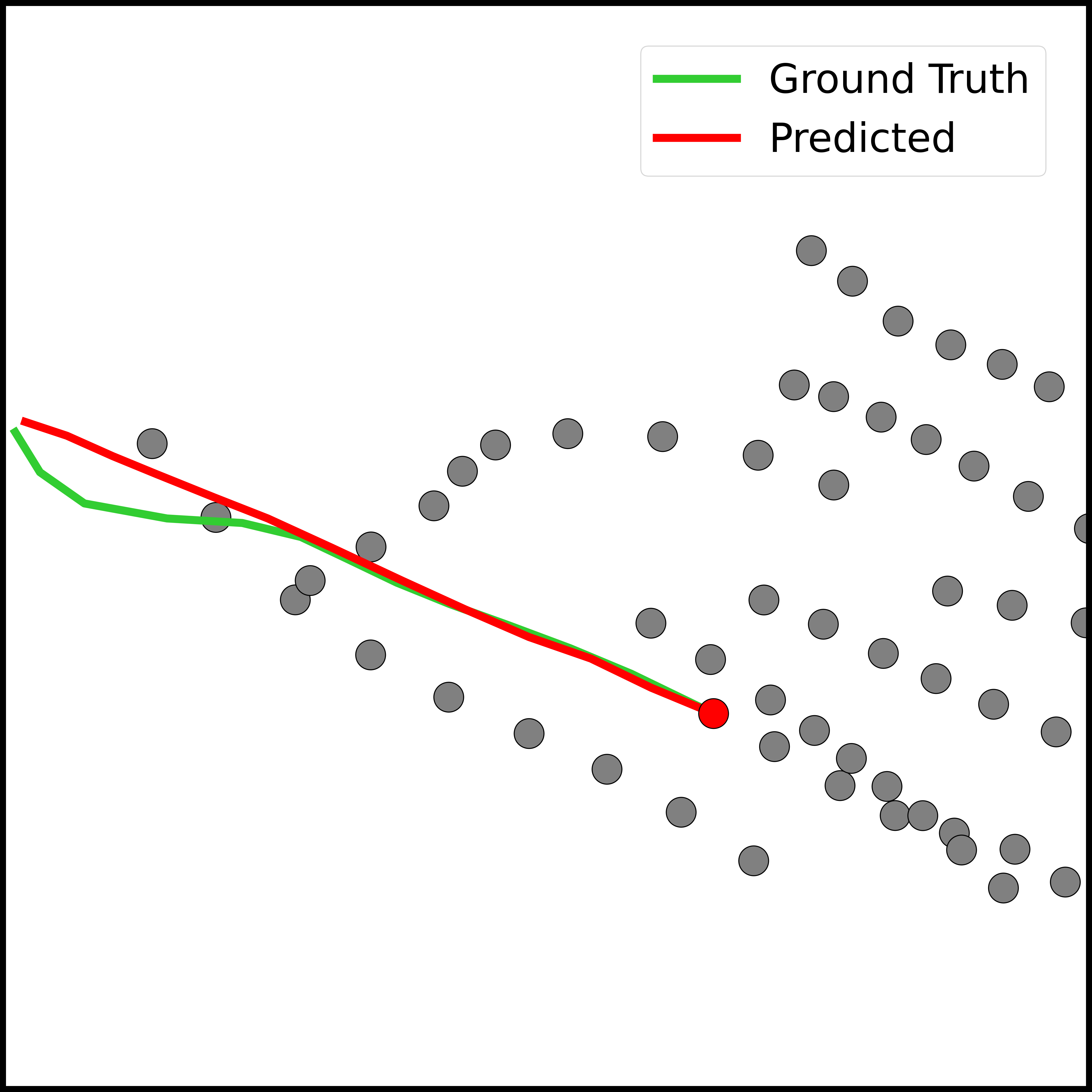} 
        (a) JRDB
    \end{center}
\end{minipage}
&
\begin{minipage}{0.5\columnwidth}
    \begin{center}
        \includegraphics[clip, width=\hsize]{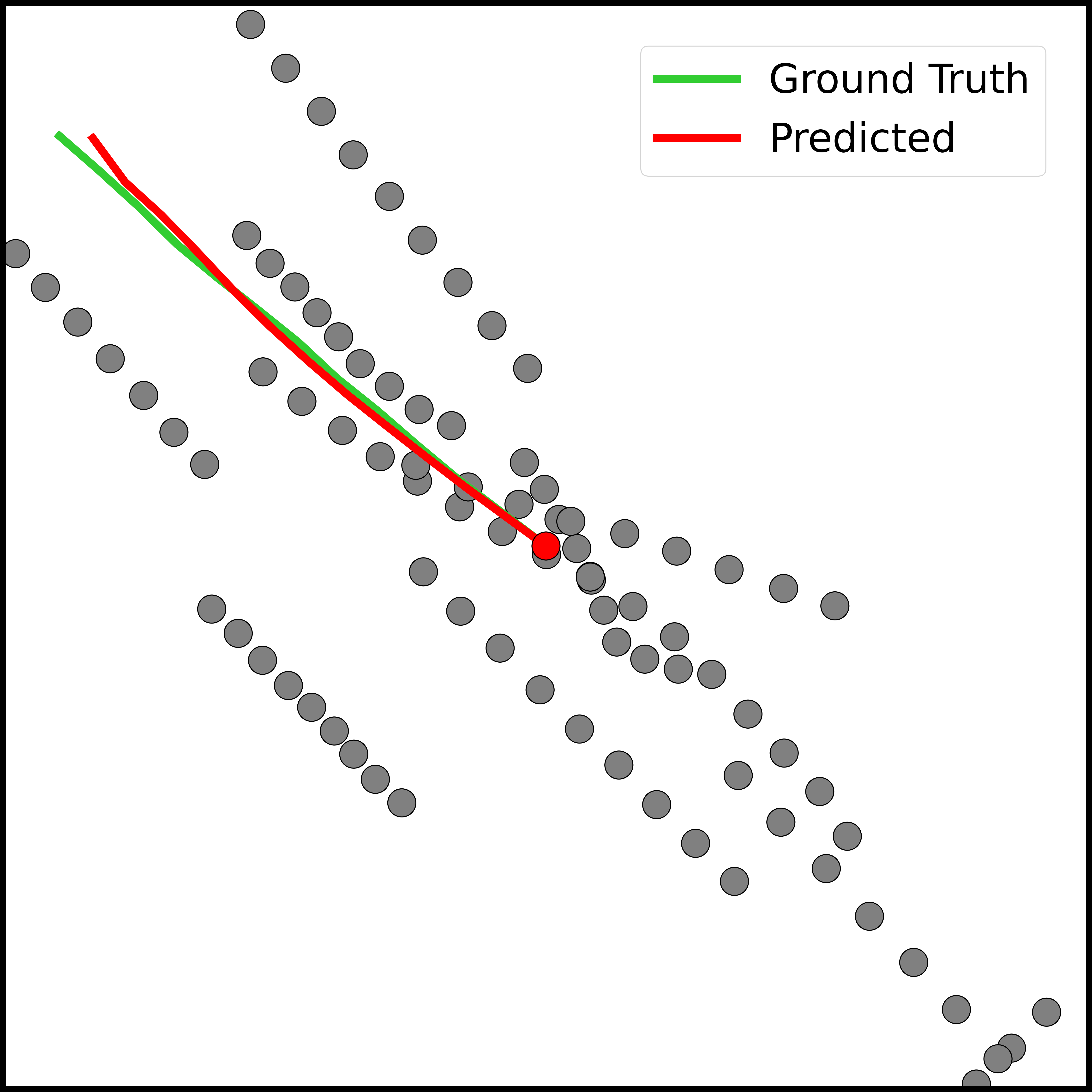}
        (b) JTA
    \end{center}
\end{minipage}
\end{tabular}}
\caption{The visualization of predicted trajectories on the JRDB and JTA. The \textcolor{red}{red} circle is the detection of the target pedestrian in the last observed frame, and the \textcolor{mygray}{gray} circles are past detections.}
\label{fig:qualitative}
\vspace{-1.em}
\end{figure}

\section{Qualitative Results}
\cref{fig:qualitative} visualizes the proposed method's predicted results and the ground-truth future trajectories, highlighting the method's accuracy in predicting future trajectories in complex scenes only using past detections as inputs.

\section{Limitation}
Similar to conventional trajectory forecasting models, our method can only predict future trajectories for pedestrians detected in the last observed frame. Ideally, predicting future paths for all pedestrians observed in any past frame, even if occluded in the last observed frame, would enhance the applicability in complex scenes. Addressing this limitation presents an interesting direction for future work.
 
\section{Conclusion}
In this paper, we introduced \textit{\paper}, a novel pedestrian trajectory forecasting framework designed to address three key limitations in conventional works. Our approach combines self-supervised pretraining on synthetic data with weakly-supervised fine-tuning on limited real-world data, reducing data collection requirements while enhancing robustness to perception errors. We proposed Det2TrajFormer, a model that leverages past detections to remain robust against tracking noise. By incorporating multiple pretext tasks during pretraining, we further improved the model's robustness and forecasting accuracy using only detection inputs. Unlike existing methods, our framework fine-tunes only using ground-truth detections, reducing person ID annotation costs. In our experiments, we thoroughly validated that our approach effectively overcomes the key limitations of existing trajectory prediction models, particularly when applied to real-world scenarios.

{
    \small
    \bibliographystyle{ieeenat_fullname}
    \bibliography{main}
}

 \clearpage
\setcounter{page}{1}
\maketitlesupplementary

\section{Methodological Comparison of Robust Trajectory Forecasting against Perception Errors} To offer a clear and concise comparison between prior works addressing perception errors and our RealTraj approach, we have summarized the key differences in~\cref{tab:survey-table}. Unlike previous methods that typically employ architectures and objectives tailored to either detection or tracking errors, RealTraj aims to develop a robust model capable of addressing both detection and tracking errors.

\begin{table}[tb]
\caption{Comparative overview of methodological features for robust trajectory forecasting against perception errors.}
\centering
\resizebox{\columnwidth}{!}{
\begin{tabular}{cccc}
\toprule
\multirow{2}{*}{Methods}  & \multicolumn{2}{c}{Detection Errors}  & Tracking Error  \\
 \cmidrule(l{2pt}r{3pt}){2-3}  \cmidrule(l{2pt}r{3pt}){4-4}
 & Miss-Detection & Error Localization & Identity Switch  \\
\hline
\cite{fujii2021two, xu2023uncovering, zhang2024oostraj, chib2024mstip, xu2024flexilength, sun2022momentary,Li2023BCDiff,monti2022howmany}  & \checkmark & -  &  -   \\
N/A  & - & \checkmark  & - \\
\cite{yu2021towardrobust, weng2022mtp, weng2022whosetrack, zhang2023fromdetection, feng2023macformer} & - & - & \checkmark  \\
\midrule
\rowcolor{Gray} 
RealTraj (Ours) & \checkmark & \checkmark & \checkmark\\
\bottomrule
\end{tabular}
}
\label{tab:survey-table}
\end{table}

\begin{figure*}[tb]
\centering
\resizebox{\textwidth}{!}{
\begin{tabular}{cccc}
\begin{minipage}{0.25\textwidth}
    \begin{center}
        \includegraphics[clip, width=\hsize]{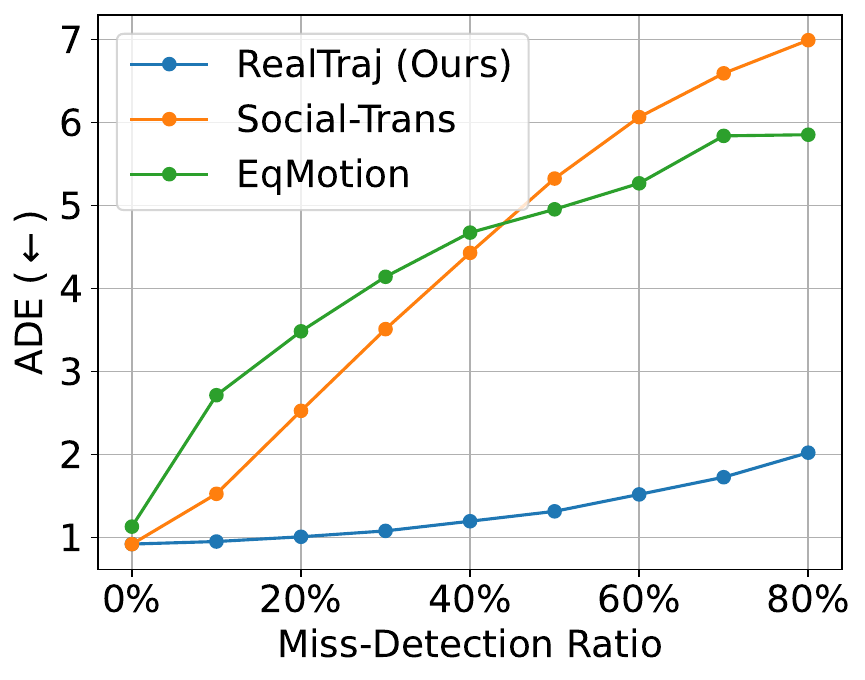} 
        (a) Miss-Detection
    \end{center}
\end{minipage}
&
\begin{minipage}{0.25\textwidth}
    \begin{center}
        \includegraphics[clip, width=\hsize]{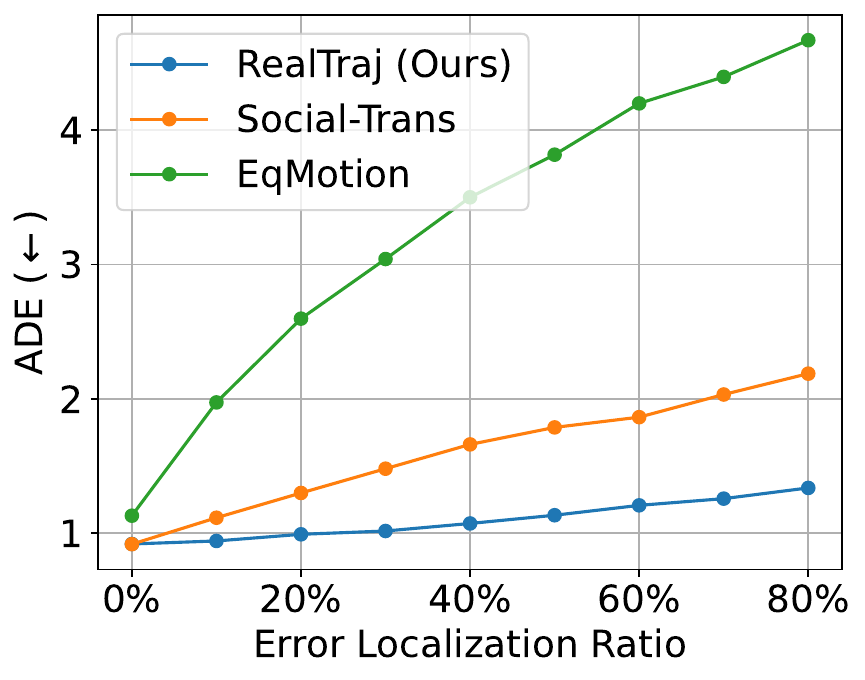} 
        (b) Error Localization
    \end{center}
\end{minipage}

&
\begin{minipage}{0.25\textwidth}
    \begin{center}
        \includegraphics[clip, width=\hsize]{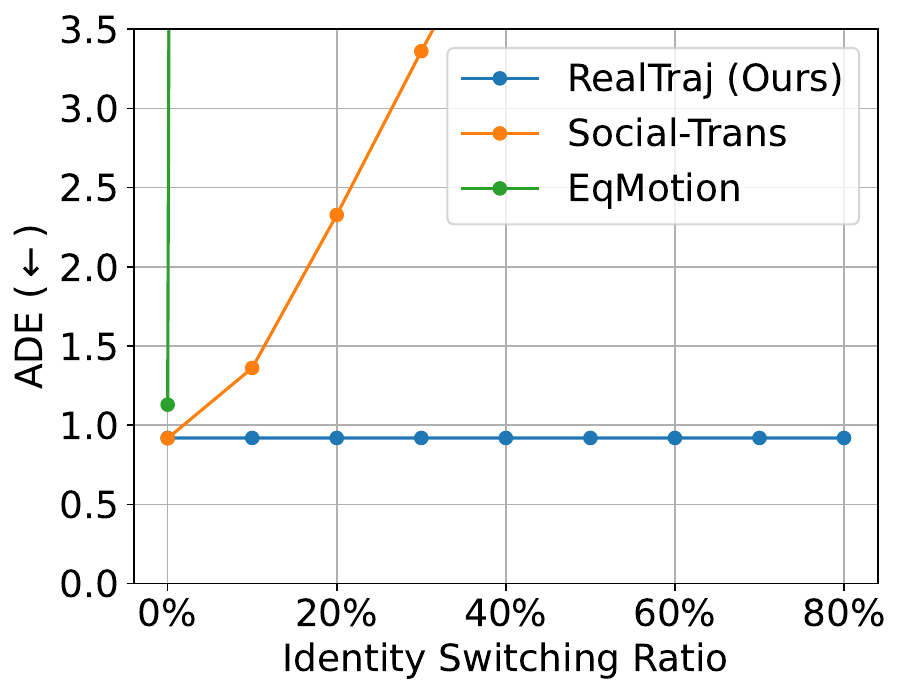} 
        (c) Identity Switch
    \end{center}
\end{minipage}

&
\begin{minipage}{0.25\textwidth}
    \begin{center}
        \includegraphics[clip, width=\hsize]{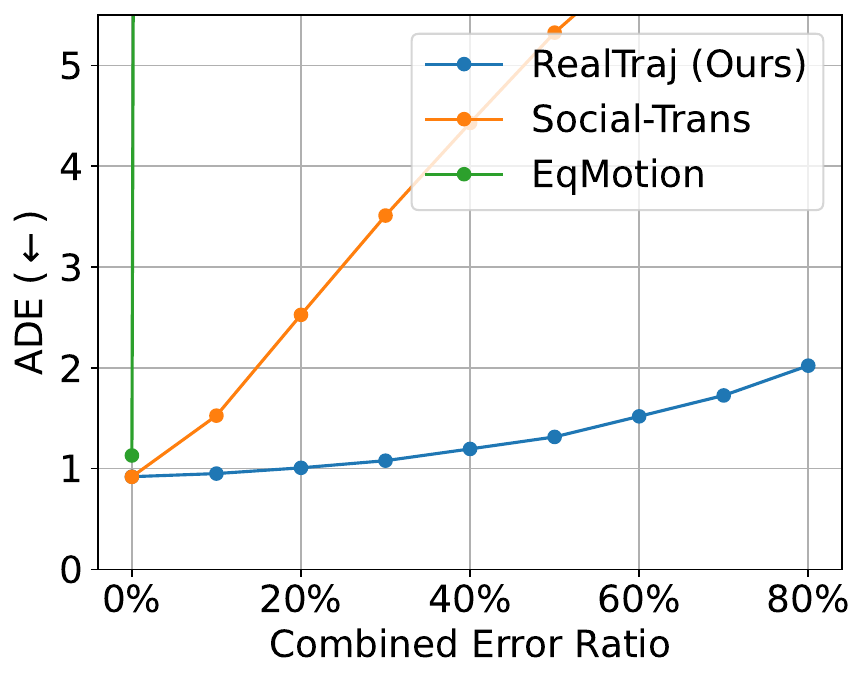} 
        (d) Combined Error
    \end{center}
\end{minipage}
\end{tabular}}
\caption{Comparison of model robustness against various error types on the JTA dataset, including detection errors (miss-detections and localization errors), tracking errors (identity switches), and their combined impact.}
\label{fig:robustness-jta}
\vspace{-1.em}
\end{figure*}

\section{Implementation Details}

As noted, we use the ORCA crowd simulator~\cite{berg2011ocra}, implemented by \cite{rempeluo2023tracepace}, to generate synthetic trajectories. Up to $40$ pedestrians are placed in a $15 \text{m} \times 15 \text{m}$ environment with up to $20$ static primitive obstacles. We generate $2000$ sequences for pretraining. 

During the first $100$ epochs of pretraining, the loss weights $\alpha$, $\beta$, and $\gamma$ are set to $1$, $0$, and $0$, respectively; for the final $100$ epochs, these weights are set to $0$, $100$, and $0.1$. A corruption ratio of $30\%$ is applied. Hyperparameters were selected through a coarse-to-fine grid search and step-by-step tuning. The model is trained with a batch size of $16$ on a single NVIDIA TITAN RTX GPU.

The model configuration for Det2TrajFormer consists of nine layers and four attention heads, with a model dimension of $d = 128$. The person ID embedding dimension $d_I$ is set to $64$. During training, we corrupt past coordinates with a corruption probability $p_c$ of $30\%$. This corruption is introduced by either simulating miss-detections or adding localization errors. Miss-detections are emulated by setting the coordinate values to zero, while localization errors are simulated by adding Gaussian noise $\epsilon \sim \mathcal{N}(0, \sigma^2)$, where the standard deviation $\sigma$ is set to $0.5$.

\section{Additional Experimental Results}
\noindent \textbf{Robustness Evaluation}
We assess the impact of detection and tracking errors introduced in the JTA dataset, as shown in~\cref{fig:robustness-jta}. The proposed method exhibits only minor performance degradation, highlighting its robustness to detection errors. Notably, RealTraj consistently outperforms all baselines across all error ratios in the JTA dataset.

\noindent \textbf{Momentary Trajectory Prediction}. We evaluate our model in a momentary trajectory prediction setting, where only two frames serve as input, and the model predicts future trajectories for the next 12 timesteps. The results in~\cref{tab:momentary} show that our model, which integrates three key limitations into a unified framework, achieves performance comparable to specialized momentary prediction approaches.

\begin{figure}[tb]
\centering
\resizebox{0.8\columnwidth}{!}{
\begin{tabular}{cc}
\begin{minipage}{\columnwidth}
    \centering
    \includegraphics[clip, width=\hsize]{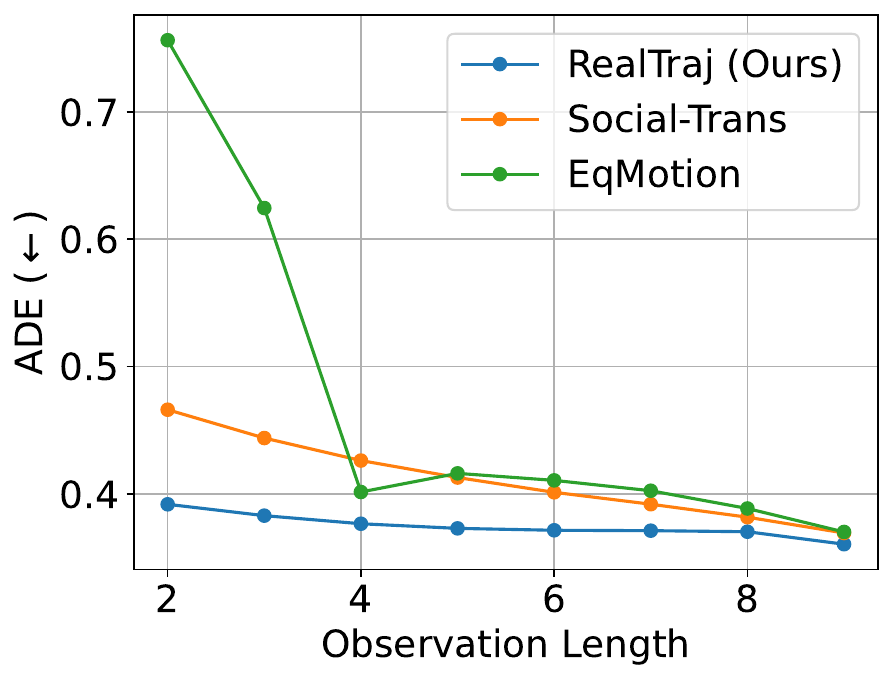} 
\end{minipage}
\end{tabular}
}
\caption{Comparisons with the current state-of-the-art method on JRDB with different observation
lengths. We report the ADE metric.The unit for ADE is meters.}
\label{fig:short-jrdb}
\vspace{-1.em}
\end{figure}

\noindent \textbf{Observation Length Shift}. To examine the model’s adaptability to observation length shift, we test it on observation lengths different from those used during training. As shown in~\cref{fig:short-jrdb}, our model consistently outperforms all baselines across varying observation lengths.

\begin{figure}[tb]
\centering
\resizebox{0.8\columnwidth}{!}{
\begin{tabular}{cc}
\begin{minipage}{\columnwidth}
    \centering
    \includegraphics[clip, width=\hsize]{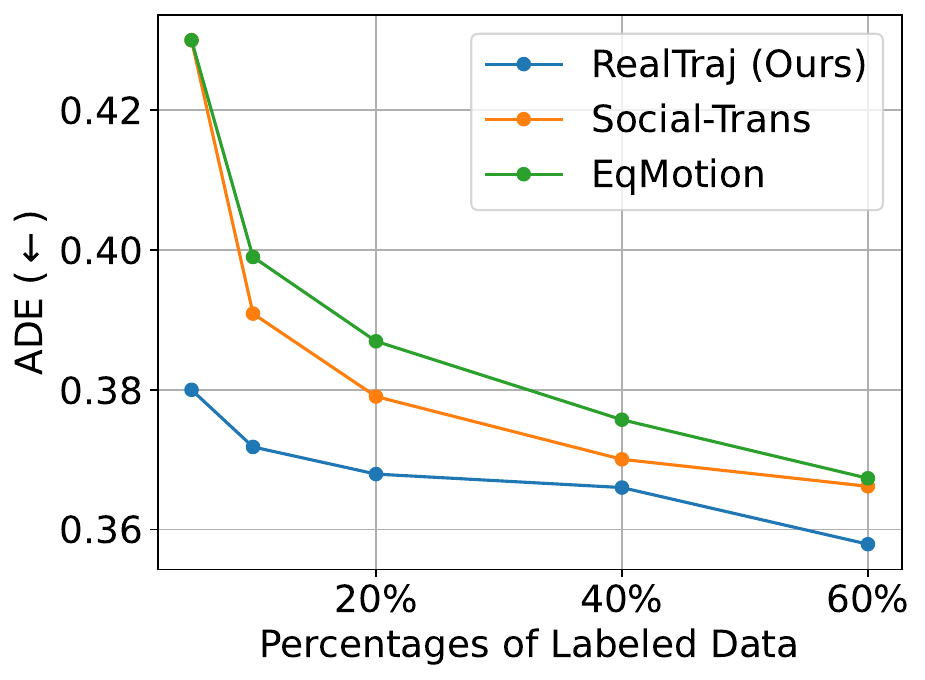} 
\end{minipage}
\end{tabular}
}

\caption{Comparisons with the current state-of-the-art method on JRDB under few-shot evaluation. We summarize the ADE metric with different percentages of labeled data. The unit for ADE is meters.}
\label{fig:additional-few-shot}
\vspace{-1.em}
\end{figure}

\noindent \textbf{Few-shot Evaluation}. In addition to the results presented in Table 1 of the main paper, we further investigated the capabilities of our model using limited real-world labeled data by randomly selecting subsets containing $5\%$, $10\%$, $20\%$, $40\%$, and $60\%$ of the dataset. The model was fine-tuned on each subset and subsequently evaluated on the test set. As shown in~\cref{fig:additional-few-shot}, our RealTraj consistently outperforms both EqMotion and Social-Trans, benefiting from pre-trained knowledge.

\noindent \textbf{Comparison of fine-tune and scratch model}.
We conducted an ablation study to evaluate whether the model benefits from pre-trained knowledge. We compared the performance of a fine-tuned model trained from a pretrained model to models trained from scratch. As shown in ~\cref{tab:finetune-vs-scratch}, the fine-tuned model consistently outperformed the scratch models by a significant margin. This suggests that pre-training with our proposed pretexts effectively enhances trajectory forecasting performance.

\begin{table}[tb]
 \caption{Comparison with the current state-of-the-art methods on ETH-UCY and SDD for momentary trajectory prediction. The best results are bolded, while the second-best results are \underline{underlined}.}
\begin{center}
\resizebox{0.9\columnwidth}{!}{
\begin{tabular}{lccccccc}
\toprule
Method & \multicolumn{2}{c}{ETH-UCY} & \multicolumn{2}{c}{SDD} \\ \cmidrule(l{2pt}r{3pt}){2-3}  
\cmidrule(l{2pt}r{3pt}){4-5}  
 & minADE$_{20}$ & minFDE$_{20}$ & minADE$_{20}$ & minFDE$_{20}$ \\ 
\toprule
MOE~\cite{sun2022momentary} & \underline{0.20} & \underline{0.41} & 8.40 & 16.08  \\
DTO~\cite{monti2022howmany} & 0.23 & 0.46 & 8.32 & \textbf{11.56} \\
BCDiff~\cite{Li2023BCDiff} & \textbf{0.19} & \textbf{0.39} & \underline{8.93} & 16.92 \\
\rowcolor{Gray}Supervised-RealTraj & 0.24 & \textbf{0.39} & \textbf{8.05} & \underline{13.32} \\ 
\bottomrule
\end{tabular}}
\end{center}
 \label{tab:momentary}
\end{table}

\begin{table}[tb]
 \caption{Comparison of fine-tune and scratch model on JRDB, JTA, ETH-UCY, and SDD. We report the ADE metric. The unit for ADE is meters. }
\begin{center}
\vspace{-0.5em}
\resizebox{1\columnwidth}{!}{
\begin{tabular}{lcccc}
\toprule
Method & JRDB & JTA & ETH-UCY & SDD \\
\toprule
Scratch & 0.78 & 1.15 & 1.05 & 17.74 \\
\rowcolor{Gray} Fine-tune (RealTraj) & \textbf{0.36} & \textbf{1.04} &\textbf{0.52} & \textbf{16.02} \\
\bottomrule
\end{tabular}}
\end{center}
 \label{tab:finetune-vs-scratch}
 \vspace{-1.5em}
\end{table}

\begin{table}[tb]
\caption{Performance results for different numbers of encoder layers in Det2TrajFormer. An encoder depth of 9 layers achieves the best balance between performance and computational efficiency.}
\begin{center}
\vspace{-0.5em}
\resizebox{0.4\columnwidth}{!}{
\begin{tabular}{lcccc}
\toprule
L & ADE & FDE \\
\toprule
3 & 0.38 & 0.76 \\
6 & 0.37 & 0.71  \\
\rowcolor{Gray} 9 & \textbf{0.36} & \textbf{0.69}\\
12 & \textbf{0.36} &  \textbf{0.69}\\
\bottomrule
\end{tabular}
}
\end{center}
 \label{tab:num-encoder-layers}
 \vspace{-1.5em}
\end{table}

\section{Effect of Number of Encoder Layers}
As shown in~\cref{tab:num-encoder-layers}, a relatively deep encoder is crucial for optimal performance. Increasing the number of encoder layers from $3$ to $9$ results in a $5.3\%$ improvement in ADE. However, adding more layers beyond this point does not yield significant gains. Therefore, we use $9$ encoder layers as our default setting, offering a favorable balance between efficiency and performance.


\end{document}